\documentclass[sigconf]{aamas}

\usepackage{array, multirow}
\usepackage{fancyvrb,fvextra}
\usepackage{balance} 

\usepackage{enumitem}       
\usepackage{float}          
\usepackage{subcaption}     
\usepackage{ragged2e}       
\usepackage{amsmath}        
\usepackage{algorithm}      
\usepackage{algpseudocode}  

\DeclareMathOperator*{\argmin}{arg\,min}

\setcopyright{ifaamas}
\acmConference[AAMAS '24]{Proc.\@ of the 23rd International Conference
on Autonomous Agents and Multiagent Systems (AAMAS 2024)}{May 6 -- 10, 2024}
{Auckland, New Zealand}{N.~Alechina, V.~Dignum, M.~Dastani, J.S.~Sichman (eds.)}
\copyrightyear{2024}
\acmYear{2024}
\acmDOI{}
\acmPrice{}
\acmISBN{}

\acmSubmissionID{1101}

\title[Cooperative Language-Guided Inverse Planning]{Pragmatic Instruction Following and Goal Assistance via Cooperative Language-Guided Inverse Planning}

\author{Tan Zhi-Xuan$^*$}
\affiliation{
  \institution{Massachusetts Institute of Technology}
  \city{}
  \country{}
}
\email{xuan@mit.edu}

\author{Lance Ying$^*$}
\affiliation{
  \institution{Harvard University}
  \city{}
  \country{}
}
\email{lanceying@seas.harvard.edu}

\author{Vikash Mansinghka}
\affiliation{
  \institution{Massachusetts Institute of Technology}
  \city{}
  \country{}
}
\email{vkm@mit.edu}

\author{Joshua B. Tenenbaum}
\affiliation{
  \institution{Massachusetts Institute of Technology}
  \city{}
  \country{}
}
\email{jbt@mit.edu}

\begin{abstract}
People often give instructions whose meaning is ambiguous without further context, expecting that their actions or goals will disambiguate their intentions. How can we build assistive agents that follow such instructions in a flexible, context-sensitive manner? This paper introduces \emph{cooperative language-guided inverse plan search} (CLIPS), a Bayesian agent architecture for pragmatic instruction following and goal assistance. Our agent assists a human by modeling them as a cooperative planner who communicates \emph{joint plans} to the assistant, then performs \emph{multimodal} Bayesian inference over the human's goal from actions and language, using large language models (LLMs) to evaluate the likelihood of an instruction given a hypothesized plan. Given this posterior, our assistant acts to minimize expected goal achievement cost, enabling it to pragmatically follow ambiguous instructions and provide effective assistance even when uncertain about the goal. We evaluate these capabilities in two cooperative planning domains (Doors, Keys \& Gems and VirtualHome), finding that CLIPS significantly outperforms GPT-4V, LLM-based literal instruction following and unimodal inverse planning in both accuracy and helpfulness, while closely matching the inferences and assistive judgments provided by human raters.
\end{abstract}


\newcommand{\BibTeX}{\rm B\kern-.05em{\sc i\kern-.025em b}\kern-.08em\TeX}

\makeatletter
\gdef\@copyrightpermission{
	\begin{minipage}{0.3\columnwidth}
		\href{https://creativecommons.org/licenses/by/4.0/}{\includegraphics[width=0.90\textwidth]{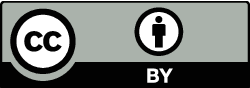}}
	\end{minipage}\hfill
	\begin{minipage}{0.7\columnwidth}
		\href{https://creativecommons.org/licenses/by/4.0/}{This work is licensed under a Creative Commons Attribution International 4.0 License.}
	\end{minipage}
	\vspace{5pt}
}
\makeatother

\begin{document}


\pagestyle{fancy}
\fancyhead{}


\maketitle 


\def\thefootnote{*}\footnotetext{Equal contribution.}\def\thefootnote{\arabic{footnote}}

\section{Introduction}

Humans act upon the world through our words. We make requests, give instructions, and communicate information, so that we can better coordinate with each other \cite{austin1975things}. In doing so, we are often parsimonious, exploiting context to convey our intentions \cite{grice1975logic}. For example, if someone says ``Can you hold that door?'', it is typically obvious which door they mean, even though nothing in the sentence distinguishes the door they just walked past from the one they are heading towards. We understand these requests because we interpret them \emph{pragmatically}, in light of the goals and actions of others. How might we build assistive machines that do the same?

\begin{figure*}[t]
\includegraphics[width=0.85\textwidth]{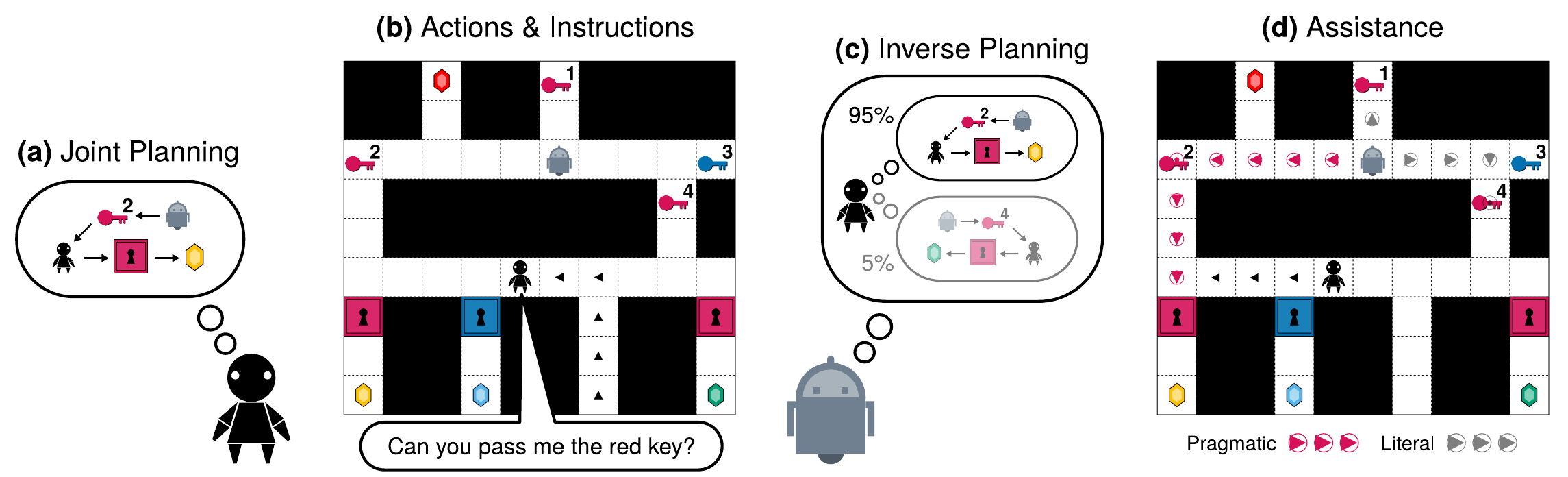}
    \caption{Overview of \emph{cooperative language-guided inverse plan search} (CLIPS). \textmd{We model a human principal as \textbf{(a)} cooperatively planning a joint policy for the human and the (robot) assistant. The human is \textbf{(b)} assumed to take actions from this joint policy while communicating planned actions as an instruction (\emph{``Can you pass me the red key?''}). Observing this, \textbf{(c)} CLIPS infers the human's goal and policy via Bayesian inverse planning. CLIPS then \textbf{(d)} acts by minimizing expected goal achievement cost, pragmatically interpreting the ambiguous instruction by picking up \emph{Key 2}. In contrast, a literal instruction follower might pick up \emph{Key 1} or \emph{Key 4}, which are also red in color.}}
    \label{fig:overview}
\end{figure*}

In this paper, we introduce \emph{cooperative language-guided inverse plan search} (CLIPS), a Bayesian architecture for pragmatic instruction following and goal assistance (Figure \ref{fig:overview}). Building upon prior work in inverse planning \cite{baker2009action,ramirez2010probabilistic, ying2023neuro}, rational speech act theory \cite{goodman2016pragmatic,fried2018unified,sumers2023reconciling}, joint intentionality \cite{tang2020bootstrapping,stacy2021modeling,wu2021too}, assistance games \cite{hadfield2016cooperative,fisac2017pragmatic}, and reward learning \cite{ramachandran2007bayesian,jeon2020reward,lin2022inferring}, CLIPS models humans as cooperative planners who communicate \emph{joint plans} as instructions. Given this model, CLIPS performs \emph{multimodal} goal inference from human actions and instructions, using a large language model (LLM) to score the likelihood of observed utterances \cite{ying2023inferring}, and computing a distribution over goals via sequential inverse planning \cite{zhi2020online}. This distribution then informs an assistive policy, which selects helpful actions under uncertainty about the human's goal.

We evaluate CLIPS on a suite of multi-step goal assistance problems in a doors-and-keys gridworld \cite{zhi2020online} and the VirtualHome domain \cite{puig2018virtualhome,puig2020watch}. In these problems, the assistant must infer the human's goal from their actions and possibly ambiguous instructions, then decide how best to help. Even when leveraging LLMs, standard instruction following methods struggle with this setting because they disregard pragmatic context \cite{ahn2022can,wan2022handmethat,singh2023progprompt}, while action-only goal inference \cite{baker2009action,zhi2020online,puig2020watch} ignores linguistic information. Multimodal LLMs have access to all information, but they fail to ground it in a coherent theory-of-mind \cite{buschoff2024visual,jin2024mmtom}. In contrast, CLIPS is able to use observed actions and inferred goals to resolve \emph{ambiguous language}, interpret \emph{joint instructions}, and correct for \emph{incomplete commands}, achieving much higher goal accuracy and cooperative efficiency than GPT-4V, LLM-based literal instruction following, and unimodal inverse planning, while correlating strongly with goal inference and assistance judgments provided by human raters.

\section{Cooperative Language-Guided Inverse Plan Search}

We formulate the setting for CLIPS as a \emph{language-augmented goal assistance game}, an extension of assistance games \cite{hadfield2016cooperative,fisac2017pragmatic} with linguistic utterances and uncertainty over goals (i.e. desired terminal states) \cite{saisubramanian2019planning}. We define this as a two-player Markov game between a human principal and an assistive agent, described by the tuple $(\mathsf{S}, \mathsf{U}, \mathsf{A}^{_h}, \mathsf{A}^{_r}, \mathsf{C}, \mathsf{G}, H, P_s, P_i)$, where $\mathsf{S}$ is the set of environment states, $\mathsf{U}$ a space of utterances $u$ that the human may use to communicate at any step, $\mathsf{A}^{_h}$ the set of human actions $a^{_h}$, $\mathsf{A}^{_r}$ the set of assistant actions $a^{_r}$, $\mathsf{C}$ a set of cost functions $C: \mathsf{S} \times \mathsf{U} \times \mathsf{A}^{_h} \times \mathsf{A}^{_r} \to \mathbb{R}$ that map state-action transitions to real numbers, $\mathsf{G} \subseteq \mathcal{P}(\mathsf{S})$ a set of possible goals $g$ where each $g \subseteq S$ is a set of (terminal) states, $H$ a horizon after which the game automatically terminates, $P_s(s'|s, a^{_r}, a^{_h})$ the environment transition distribution, and $P_i(s_i, C, g)$ a distribution over initial states of the game. As in regular assistance games, the human knows (i.e. observes) the \emph{true} cost function $C$ and goal $g$ sampled from $P_i$, but the assistant only observes the initial environment $s_i$. Thus, the assistant has to \emph{infer} the true goal $g$.

Assistance games admit cooperative equilibria, but they are intractable to compute in general \cite{fisac2017pragmatic}, and also assume more game-theoretic optimality from humans than may be warranted \cite{milli2020literal}. As such, our goal is to \emph{approximately} solve the assistance game under reasonable modeling assumptions about how humans cooperate and communicate. By fixing a human model, the game becomes a partially observable Markov decision process (POMDP) from the assistant's perspective \cite{doshi2009monte}, which can be solved through approximate methods \cite{nguyen2011capir,macindoe2012pomcop}. We describe this human model, then show how an assistant can perform Bayesian inference over such a model, using the acquired information to better assist humans.

\subsection{Modeling cooperative action and communication}

\begin{figure*}[t]
    \centering
    \begin{subfigure}[b]{0.45\textwidth}
    \centering
    \includegraphics[width=0.85\columnwidth]{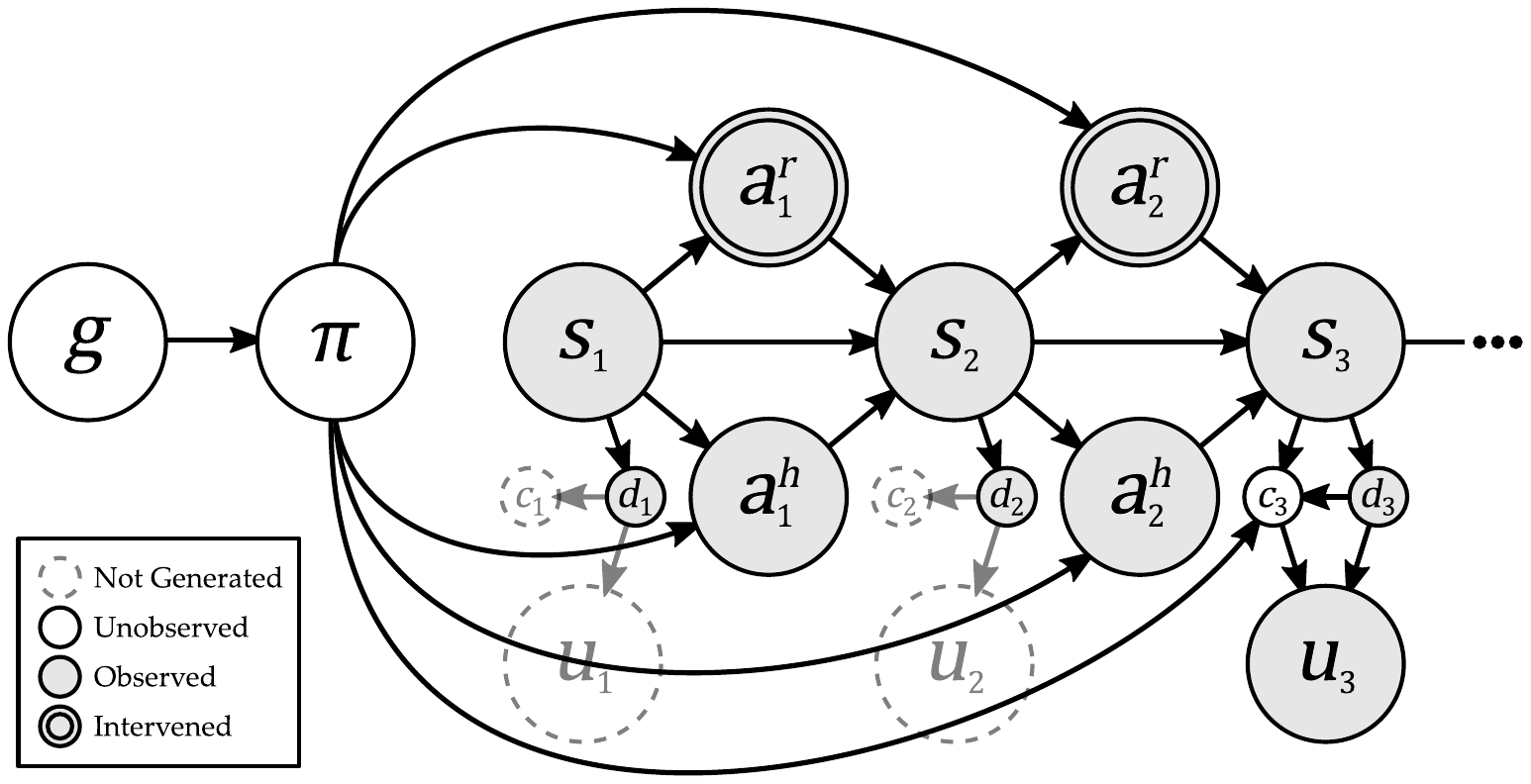}
    \caption{\textmd{One graphical realization of our Bayesian model.}}
    \label{fig:graphical-model}
    \vspace{2pt}
    \algrenewcommand\algorithmicprocedure{\textbf{model}}
    \begin{algorithmic}
        \Procedure{clips-model}{$s_1, G, T$}
        \State $g \sim \textsc{goal-prior}(G)$
        \State $\pi \sim \textsc{policy-init}(g, s_1)$
        \For{$t \in [1, T]$}
            \State $\pi \gets \textsc{rths-policy-update}(\pi, g, s_t)$
            \State $a^{_h}_t, a^{_r}_t \sim \textsc{boltzmann-dist}(s_t, \pi)$
            \State $u_t, c_t, d_t \sim \textsc{utterance-model}(s_t, \pi)$
            \State $s_{t+1} \sim \textsc{transition}(s_t, a^{_h}_t, a^{_r}_t)$
        \EndFor
    \EndProcedure
    \end{algorithmic}
    \caption{\textmd{CLIPS agent-environment model as a probabilistic program.}}
    \label{fig:clips-model}
    \end{subfigure}
    \hspace{6pt}
    \begin{subfigure}[b]{0.40\textwidth} 
    \algrenewcommand\algorithmicprocedure{\textbf{model}}
    \begin{algorithmic}
        \Procedure{utterance-model}{$s_t, \pi$}
        \State \textbf{parameters}: $p_\text{speak}, H, K, \mathcal{E}$
        \State $d_t \sim \textsc{bernoulli}(p_\text{speak})$
        \If{$d_t = \textsc{true}$}
            \State $a_{t:t+H} \gets \textsc{rollout-policy}(s_t, \pi, H)$
            \State $\alpha_{1:m} \gets \textsc{extract-salient-actions}(a_{t:t+H})$
            \State $c_t \sim \textsc{rand-subset}(\alpha_{1:m}, K)$
            \State $u_t \sim \textsc{language-model}(c_t, \mathcal{E})$
        \Else
            \State $c_t, u_t \gets \emptyset, \texttt{""}$
        \EndIf
        \EndProcedure
    \end{algorithmic}
    \caption{\textmd{Utterance model $P(u_t, c_t, d_t| \pi)$ as a probabilistic program.}}
    \label{fig:utterance-model}

    \vspace{8pt}

    \scriptsize
    \begin{tabular}{rp{0.79\columnwidth}}
    \textbf{Command:} & \RaggedRight{\texttt{(handover you me key1) where (iscolor key1 blue)}} \\
    \textbf{Utterance:} & Hand me the blue key. \\
    \textbf{Command:} & \RaggedRight{\texttt{(unlock you key1 door1)}} \\
    \textbf{Utterance:} & Can you unlock the door for me? \\
    \textbf{Command:} & \RaggedRight{\texttt{(pickup me key1) (unlock you key2 door1) where (iscolor key1 blue) (iscolor door1 green) }} \\
    \textbf{Utterance:} & I'm getting the blue key, can you open the green door? \\
    \textbf{Command:} & \RaggedRight{\texttt{(handover you me key1) (handover you me key2) where (iscolor key1 green) (iscolor key2 red)}} \\
    \textbf{Utterance:} & Can you pass me the green and red keys? \\
    \end{tabular}
    \caption{\textmd{Paired examples $\mathcal{E}$ of commands $c_t$ and utterances $u_t$.}}
    \label{fig:utterance-examples}
    \end{subfigure}
    
    \caption{Model architecture. \textmd{In CLIPS, we model the human as a cooperative planner who computes a joint policy $\pi$ for a goal $g \in G$.
    The policy $\pi$ dictates the human's and assistant's actions $a^{_h}_t, a^{_r}_t$ at each state $s_t$, as well as the command $c_t$ and utterance $u_t$ that the human may decide $d_t$ to communicate at step $t$. One realization of this process is depicted in \textbf{(a)}, showing a case where an utterance $u_3$ is only made at $t=3$. We implement this process as probabilistic program, shown in \textbf{(b)}. Utterance generation is modeled by the subroutine in \textbf{(c)}, which summarizes salient actions from policy $\pi$ as a command $c_t$, then samples an utterance $u_t$ using a (large) language model prompted with $c_t$ and \textbf{(d)} a list of few-shot examples $\mathcal{E}$ demonstrating how commands are translated into natural language.}}
    \label{fig:clips-model-full}
\end{figure*}

When humans cooperate, we direct our actions towards a shared goal while expecting that others will do the same. This capacity for \emph{joint intentionality} \cite{tomasello2007shared} allows us to cooperate rapidly and flexibly while avoiding nested reasoning about each other's minds \cite{tang2020bootstrapping,wu2021too}. In CLIPS, we exploit this aspect of human cognition by modeling the human principal as a \emph{cooperative planner}, who computes a \emph{joint policy} $\pi$ for both agents to achieve the goal $g$. The human follows $\pi$ by taking an action $a^{_h}_t$ at each step $t$, and assumes the assistant will take an action $a^{_r}_t$ accordingly. These actions lead to a change in the state $s_t$. The human may also speak at any time $t$ (with the decision denoted by $d_t$), issuing a command $c_t$ that summarizes their intended policy $\pi$. This command is then rendered in natural language as an utterance $u_t$. The overall generative process is shown in Figures \ref{fig:clips-model-full}a--b, and can be summarized as follows:
\begin{alignat}{3}
&\textit{Goal Prior:} \qquad &g &\sim P(g) \label{eq:goal-prior} \\
&\textit{Joint Planning:} \qquad &\pi &\sim P(\pi|g) \label{eq:joint-planning} \\
&\textit{Action Selection:} \qquad &a^{_h}_t, a^{_r}_t &\sim P(a^{_h}_t, a^{_r}_t | s_t, \pi) \label{eq:action-selection} \\
&\textit{Utterance Generation:} \qquad &u_t, c_t, d_t &\sim P(u_t, c_t, d_t| s_t, \pi) \label{eq:utterance-generation} \\
&\textit{State Transition:} \qquad &s_{t+1} &\sim P(s_{t+1} | s_t, a^{_h}_t, a^{_r}_t) \label{eq:state-transition}
\end{alignat}

$P(g)$ is the assistant's prior over the human's goal $g \in \mathsf{G}$, which we take to be uniform. We can also fold in uncertainty over the human's cost function $C \in \mathsf{C}$ into this prior, treating $g$ as a \emph{specification} that includes both the goal condition and action costs.

To model joint planning, we assume that the human computes a Boltzmann policy $\pi$ for goal $g$, which defines an approximately rational distribution over joint actions $a^{_h}_t, a^{_r}_t$:
\begin{equation}
    \pi(a^{_h}_t, a^{_r}_t | s_t) = \frac{\exp \left(-\beta \hat Q_{\pi}(s_t, a^{_h}_t, a^{_r}_t)\right)}{\sum_{\tilde{a}^{_h}_t, \tilde{a}^{_r}_t} \exp\left(-\beta \hat Q_{\pi}(s_t, \tilde{a}^{_h}_t, \tilde{a}^{_r}_t)\right)}
    \label{eq:boltzmann}
\end{equation}
Here, $\beta$ is a rationality parameter controlling the optimality of the policy (higher $\beta$ is more optimal), over which we may place a prior $P(\beta)$, and $\hat Q_{\pi}(s_t, a^{_h}_t, a^{_r}_t)$ is an estimate of the expected cost of reaching the goal $g$ by taking actions $a^{_h}_t, a^{_r}_t$ at state $s_t$. To estimate $\hat Q_{\pi}$ efficiently, we extend prior work in online goal inference \cite{zhi2020online}, using model-based planning to compute policies on-the-fly. In particular, we use real-time heuristic search (RTHS) as an anytime planner \cite{korf1990real,barto1995learning,koenig2006real}, which estimates $Q$-values in a neighborhood around the current state $s_t$ via search (\textsc{rths-policy-update} in Fig. \ref{fig:clips-model}), while using previously estimated $Q$-values to guide future searches. Unlike earlier methods for inverse planning, this avoids the intractability of estimating $Q$ for all states and actions \cite{ramachandran2007bayesian,ziebart2008maximum}.\footnote{As an additional simplification, we assume that the human and the assistant take turns while acting (i.e. $a^{_r}_t$ is a no-op when $a^{_h}_t$ is not, and vice versa). This reduces the branching factor while planning, but preserves the optimal solution \cite{boutilier1996planning}.}

Having computed the policy $\pi$, we model human communication as an approximately \emph{pragmatic} process. Since the human wants to achieve their goal, they are likely to convey information about the policy they have in mind. However, long utterances are costly, and so the human is likely to convey only a salient summary of the policy, reducing the cost of communication. We capture these aspects of cooperative communication through a structured generation procedure that leverages the linguistic competence of LLMs:

\begin{enumerate}
    \item At each step $t$, the human decides whether to communicate ($d_t$) with probability $p_\text{speak}$. If $d_t$ is true, we rollout the policy $\pi$ for $H$ steps, producing a series of planned actions $a_{t:t+H}$.
    \item The planned actions $a_{t:t+L}$ are filtered down to a domain-specific set of \emph{salient} actions $\alpha_{1:m}$, and a random subset of up to $K$ such actions are incorporated into a command $c_t$.
    \item  Finally, the command $c_t$ is translated into a natural language utterance $u_t$ using an LLM prompted with few-shot examples (Figure \ref{fig:utterance-examples}). This defines a likelihood function $P(u_t|c_t)$.
\end{enumerate}

This procedure is implemented as the probabilistic program shown in Figure \ref{fig:utterance-model}. While it does not capture all ways in which a pragmatic speaker might prefer some utterances over others, key features of pragmatic speech are modeled: Utterances are about plans, and hence practically useful, but are also restricted to small ($\leq K$) but salient subsets of those plans, ensuring they are informative without being costly. Furthermore, using an LLM as an utterance model $P(u_t|c_t)$ provides broad coverage, enabling CLIPS to handle much richer utterances than traditional pragmatic models \cite{goodman2016pragmatic}.

\subsection{Goal inference via inverse planning \\ from actions and instructions}
\label{sec:goal-inference}

Using this probabilistic model, our assistant can infer a posterior distribution over the human's goals $g$ and policies $\pi$ given a sequence of observed actions $a^{_h}_{1:T}$, states $s_{1:T}$ utterance decisions $d_{1:T}$, and utterances $u_{1:T}$ (where $u_t$ is the empty string whenever $d_t = \textsc{false}$). Some care must be taken, however. Since the assistant is also taking actions $a^{_r}_{1:T}$ alongside the human principal, it is tempting to infer the human's goal by conditioning on \emph{both} agents' actions:
\begin{align}
    P(&g, \pi|s_{1:T}, d_{1:T}, u_{1:T}, a^{_h}_{1:T}, a^{_r}_{1:T}) \nonumber \\
    &\propto P(g, \pi, s_{1:T}, d_{1:T}, u_{1:T}, a^{_h}_{1:T}, a^{_r}_{1:T}) \label{eq:posterior-external} \\
    &= P(g, \pi) \textstyle{\prod}_{t=1}^{T}  P(s_t | s_{t-1}, a^{_h,_r}_{t-1}) P(u_t, d_t | s_t, \pi) \underline{P(a^{_h}_t, a^{_r}_t|s_t, \pi)} \nonumber
\end{align}

For an \emph{external} observer, this is the appropriate distribution to compute \cite{shum2019theory, ying2023inferring}. However,  prior work on cooperative inference also computes Equation \ref{eq:posterior-external} when observers are \emph{internal} to the environment \cite{tang2020bootstrapping,wu2021too}, which can lead to pathologies. For example, the assistant might condition on the fact that it has not moved so far ($a^{_r}_{1:T}$ are no-ops), leading it to infer that the human is pursuing a goal $g$ where the assistant's help is unnecessary. To address this, it is crucial to recognize that the assistant is \emph{intervening} upon the environment through its actions (Figure \ref{fig:graphical-model}, double-lined nodes).\footnote{Alternatively, it is enough to note the assistant's actions are selected according to \emph{different policy} $\pi'$ than the inferred joint policy $\pi$ computed by the human. Thus it is \emph{safer} for the assistant not to update its beliefs as if its actions come from $\pi$.} As such, we should detach the evidential connection between the assistant's actions $a^{_r}_{1:T}$ and any causal ancestors, which we denote using Pearl's \textbf{do}-operator \cite{pearl2012calculus}:
\begin{align}
    P(&g, \pi|s_{1:T}, d_{1:T}, u_{1:T}, a^{_h}_{1:T}, \mathbf{do}(a^{_r}_{1:T})) \nonumber \\
    &\propto P(g, \pi, s_{1:T}, d_{1:T}, u_{1:T}, a^{_h}_{1:T}, \mathbf{do}(a^{_r}_{1:T})) \label{eq:posterior-internal} \\
    &= P(g, \pi) \textstyle{\prod}_{t=1}^{T}  P(s_t | s_{t-1}, a^{_h,_r}_{t-1}) P(u_t, d_t | s_t, \pi) \underline{P(a^{_h}_t|s_t, \pi)} \nonumber
\end{align}
The difference between Equations \ref{eq:posterior-external} and \ref{eq:posterior-internal} lies in the final underlined term: Whereas an external observer reweights its beliefs by incorporating the likelihood of both agents' actions $P(a^{_h}_t, a^{_r}_t|s_t, \pi)$, our assistive agent should only incorporate the likelihood of the human's action $P(a^{_h}_t |s_t, \pi)$.

We compute the distribution in Equation \ref{eq:posterior-internal} sequentially, as shown in Algorithm \ref{alg:clips-inference}. An initial belief $b_0$ is returned by \textsc{belief-init}, which generates a set of $N$ weighted samples by either sampling or enumerating over the goal prior $P(g)$ and policy distribution $P(\pi | g)$ (accounting e.g. for uncertainty over the rationality parameter $\beta$). Then at each step $t$, the previous belief $b_{t-1}$ is updated by adjusting the weight $w^i$ associated with each goal $g^i$ and policy $\pi^i$ sample. To do this, \textsc{belief-update} first refines the $Q$-value estimates that define the policy $\pi^i$ by running more iterations of RTHS (\textsc{rths-policy-update} in L9). The weights are then updated to reflect the new state $s_t$ and decision to speak $d_t$ (L10-11). If $d_t$ is true (i.e. $u_t$ is observed), we also marginalize over all commands $c_t $ that could have generated $u_t$ from $\pi$, and update $w^i$ with the resulting mixture likelihood (L13-14). Finally, we condition on the human's action $a^{h}_t$, and return the resulting sample collection (L16-18).

If we enumerate over all possible goals and policy configurations in \textsc{belief-init}, this is an \emph{exact} Bayesian inference algorithm. In environments where there are too many hypotheses for this to be efficient, Algorithm \ref{alg:clips-inference} can readily be extended to a sequential Monte Carlo algorithm \cite{del2006sequential,lew2023smcp3} in the style of sequential inverse plan search \cite{zhi2020online}. For our experiments, however, enumeration is feasible, and we opt for this approach to avoid variance from sampling.

\subsection{Pragmatic instruction following as \\ goal assistance under uncertainty}

With Algorithm \ref{alg:clips-inference}, our assistive agent is able to infer a distribution over the human's goal $g$ and policy $\pi$ at each step $t$. 
How should it use this information to act, especially if the goal $g$ remains uncertain? In keeping with our Bayesian approach, we minimize \emph{expected} cost, acting to help the human achieve their likely goal as quickly as possible \cite{shah2020planning}. Since each inferred policy $\pi^i$ corresponds to a $\hat Q_{\pi^i}$-value function, the assistant can select actions by minimizing the expected $\hat Q_{\pi^i}$ value given its uncertainty over $\pi^i$:
\begin{equation}
    a^{_r*}_t = \textstyle{\argmin}_{a^{_r}_t} \mathbb{E}_{\pi^i}[\hat Q_{\pi^i}(s_t, a^{_h}_t, a^{_r}_t)]
\end{equation}

We implement this assistance policy with Algorithm \ref{alg:clips-assistance}, which effectively solves the $Q_\text{MDP}$ approximation of the assistive POMDP \cite{littman1995learning,hauskrecht2000value,macindoe2012pomcop}. At each step $t$, we assume the assistant has already updated its belief $b_{t-1}$ from the previous step $t-1$, and also observes the human's action \footnote{This assumption is natural in our turn-based setting. When actions are simultaneous the assistant can minimize expectation over both policies $\pi^i$ and human actions $a^{_h}_t$.}. We then iterate over all samples $(w^i, g^i, \pi^i)$, update the policies $\pi^i$ if necessary, and perform a weighted sum of $\hat Q_{\pi^i}$-values for all assistant actions $a^{_r}_t$ according to their inferred weights $w^i$. This produces a vector of \emph{assistive} $Q$-values, which we minimize over to compute the best assistive action $a^{_r*}_t$.

\begin{algorithm}[t]
\caption{CLIPS belief initialization and update}
\label{alg:clips-inference}
\begin{algorithmic}[1]
\Procedure{belief-init}{$N$}
    \State $w^i \gets 0$;\quad$g^i \sim P(g)$;\quad$\pi^i \sim P(\pi | g^i)$ \textbf{for} $i \in [1,N]$
    \State $b_0 \gets \{(w^i, g^i, \pi^i)\}_{i=1}^N$; \Return $b_0$
\EndProcedure
\\
\Procedure{belief-update}{$b_{t-1}, s_{t-1:t}, d_t, u_t, a^{_h}_{t-1:t}, a^{_r}_{t-1}$}
    \State $\{(w^i, g^i, \pi^i)\}_{i=1}^N \gets b_{t-1}$ 
    \For{$i \in [1, N]$}
        \State $\pi^i \gets \textsc{rths-policy-update}(\pi^i, g^i, s_t)$
        \State $w^i \gets w^i \cdot P(s_t|s_{t-1}, a^{_h}_{t-1}, a^{_r}_{t-1})$ 
        \State $w^i \gets w^i \cdot P(d_t|s_t, \pi)$ 
        \If{$d_t = \textsc{true}$} 
            \State $P(u_t |s_t, \pi) \gets \sum_{c_t} P(u_t | c_t) P(c_t |s_t, \pi)$
            \State $w^i \gets w^i \cdot P(u_t |s_t, \pi)$
        \EndIf
        \State $w^i \gets w^i \cdot P(a^{_h}_t|s_t, \pi)$
    \EndFor
    \State $w^i \gets w^i / \sum_{j=1}^N w_j$
    \State $b_t \gets \{(w^i, g^i, \pi^i)\}_{i=1}^N$; \Return $b_t$
\EndProcedure
\end{algorithmic}
\end{algorithm}

\begin{algorithm}[t]
\caption{CLIPS $Q_\text{MDP}$ assistance policy}
\label{alg:clips-assistance}
\begin{algorithmic}[1]
\Procedure{assistance-policy}{$b_{t-1}$, $s_t$, $a^{_h}_t$}
    \For{$a^{_r}_t \in \mathcal{A}^{_r}(s_t)$}
        \State $\hat Q_\text{assist}(s_t, a^{_h}_t, a^{_r}_t) \gets 0$
        \For{$(w^i, g^i, \pi^i) \in b_{t-1}$}
            \State $\pi^i \gets \textsc{rths-policy-update}(\pi^i, g^i, s_t)$
            \State $\hat Q_\text{assist}(s_t, a^{_h}_t, a^{_r}_t) \overset{_+}{\gets} w^i \cdot \hat Q_{\pi^i}(s_t, a^{_h}_t, a^{_r}_t)$
        \EndFor
    \EndFor
    \State $a^{_r*}_t \gets \arg\min_{a^{_r}_t} Q_\text{assist}(s_t, a^{_h}_t, a^{_r}_t)$
    \State \Return $a^{_r*}_t$
\EndProcedure
\end{algorithmic}
\end{algorithm}

\section{Experiments}

\begin{figure}
    \centering    \includegraphics[width=\columnwidth]{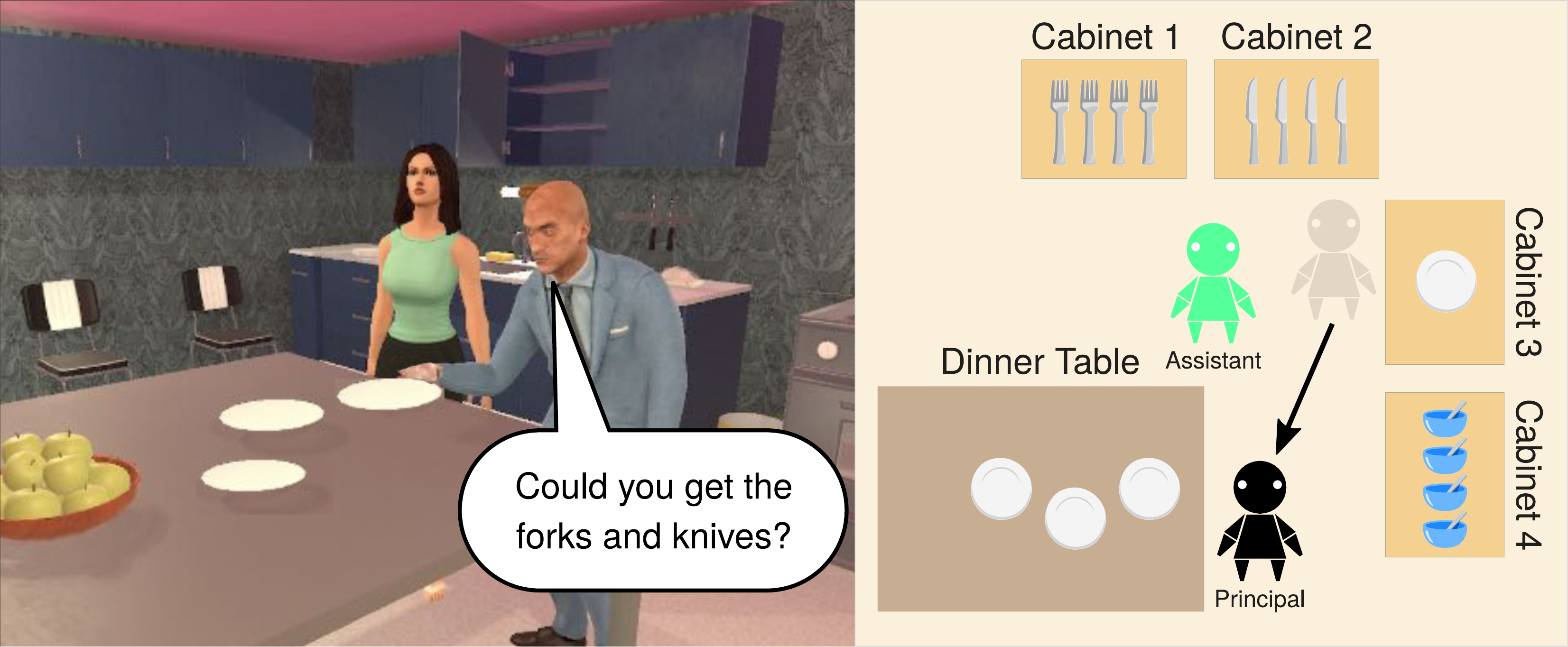}
    \caption{Example goal assistance problem in VirtualHome, \textmd{where the principal and assistant collaborate to set the dinner table. The principal places three plates on the table, then says \emph{``Could you get the forks and knives?''}. A pragmatic assistant has to infer the number of forks and knives from context (in this case, three each).}}
    \label{fig:vh-example}
\end{figure}

We evaluate our method against a variety of baselines in two domains: a cooperative gridworld puzzle called \textbf{multi-agent Doors, Keys \& Gems} (mDKG) originally developed in \cite{zhi2020online,ying2023inferring}, and \textbf{VirtualHome} (VH), a virtual household simulator \cite{puig2018virtualhome,puig2020watch}. In both domains, a human principal and a (robot) assistant cooperate on multi-step tasks to accomplish the human's goals. In mDKG, the goals are four colored gems which are often secured behind doors. Doors can be unlocked by keys of the same color, and the assistant can help by collecting keys or unlocking doors (Figure \ref{fig:overview}). In VirtualHome, the goals are 6--12 household tasks, including setting up tables and preparing ingredients (Figure \ref{fig:vh-example}). Both domains admit encodings in the Planning Domain Definition Language (PDDL) \cite{mcdermott1998pddl} with some extensions \cite{zhi2022pddl}, which we use as the environment representation for our planning and inverse planning algorithms. 

To systematically test the ability of assistive agents to pragmatically follow a human's instructions and assist with their goals, we developed a dataset of \emph{goal assistance problems} for each domain (30 problems in mDKG, 25 problems in VH). In each problem, the goal is initially unknown to the assistant. While taking a sequence of actions $a^{_h}_{1:T}$, the human communicates one or more instructions $u_{1:T}$ to the assistant. The assistant then has to make an inference about the human's goal $g$, and chooses actions to best assist them. We designed our problems to cover a range of scenarios inspired by human communication we observed in exploratory studies, varying the information that can be deciphered from the human's actions or instructions (see Figure \ref{fig:dkg-examples} and Supplementary Information). 
In mDKG, we restricted the human to only one instruction $u_T$, since more information would render goal inference trivial. In VH, however, the goal space was larger, allowing us to construct 10 problems with multiple utterances such as  \emph{``Can you get the plates?''} followed by \emph{``Bring the bowls too!''}

\subsection{Model configuration}

In our experiments, we configured CLIPS to compute $Q$-values using the real-time adaptive A* variant of RTHS \cite{koenig2006real}, and a $\operatorname{Gamma}$ prior over the rationality parameter $\beta$. For the utterance model, we used either OpenAI's 6.7B \texttt{curie} model \cite{brown2020language} or the \texttt{davinci-002} as our LLMs, due to the more diffuse probabilities they provided as base models. For goal assistance, we ran Algorithm \ref{alg:clips-assistance} in offline mode, fixing the inferred posterior after observing $a_{1:T}^h$ and $u_{1:T}$, then selecting actions by minimizing expected $Q$-values with respect to the fixed posterior. This was evaluated against a simulated human agent which followed a joint policy to the true goal, unless it was apparent that the assistant was not doing the same. While Algorithm \ref{alg:clips-assistance} can also be run in online mode, we fixed the posterior to better match the information that our human raters were provided. 

\subsection{Baselines}

To evaluate the benefit of pragmatic instruction following over either a literal interpretation of instructions, or goal inference from only one modality, we included the following baselines:

\subsubsection{Unimodal Inverse Planning}
We implemented action-only and language-only inverse planning (IP) baselines as ablations of CLIPS, where the action-only baseline is similar to sequential inverse plan search \cite{zhi2020online} for inference and Watch-and-Help \cite{puig2020watch} for assistance. In these baselines, the assistive agent uses the same belief update shown in Algorithm \ref{alg:clips-inference}, except that it conditions on only actions $a^{_h}_{1:t}$ or only the utterance $u_t$. In cases with no actions, CLIPS and the language-only baseline are equivalent.

\subsubsection{LLM-Based Literal Listener}
The literal listener baseline interprets the instruction $u_t$ in state $s_t$ without further information about the human's actions or goals. To implement this baseline, we adapt the utterance model in Figure \ref{fig:utterance-model} to sample from the space of all assistant-directed commands $c_t$ that are possible in state $s_t$. In particular, we enumerate over all assistant actions $\mathcal{A}^{_r}$, and select a subset of up to $K$ salient actions to form a command $c_t$. This defines a distribution over commands $P(c_t | s_t)$ which \emph{does not} depend the principal's policy $\pi$ or goal $g$. Given the command $c_t$, we use an LLM (\texttt{text-davinci-002} for mDKG, \texttt{davinci-002} for VH) as an utterance likelihood $P(u_t | c_t)$. Assuming a uniform prior over commands, we can perform enumerative inference to compute the posterior $P(c_t | u_t, s_t)$ given an instruction $u_t$ in state $s_t$.

Given a command $c_t$, the assistant still needs to generate an assistive plan.
We did this in two ways: The \emph{naive} literal assistant interprets a command $c_t$ by randomly selecting one concrete grounding (e.g. \emph{``Can you pick up the red key?''} could mean picking up the red key closest to the human or some other key), then planning to achieve that concrete goal. In contrast, the \emph{efficient} literal assistant tries to directly satisfy the command $c_t$ in the most efficient way\footnote{Note that the naive/efficient distinction was not meaningful in the VirtualHome problems, where utterance ambiguity was not due to grounding ambiguity.} (e.g. \emph{``Could you unlock the blue door?''} is satisfied by unlocking the blue door closest to the assistant). For both variants, the assistant can either satisfy the most likely command $c_t^*$, or sample a command from the full distribution $P(c_t|u_t, s_t)$. We report results for the latter after averaging over 10 samples, using systematic sampling to reduce variance.

\subsubsection{Multimodal LLM (GPT-4V)}

For the mDKG domain, we used GPT-4 with Vision (GPT-4V) \cite{achiam2023gpt} as a purely neural baseline, prompting it with the same set of rules and instructions that we provided to our human raters, along with the final frame of each animated visual that we showed to humans (similar to Figure \ref{fig:overview}(b)). This allowed us to probe the degree to which multimodal LLMs are capable of intuitive pragmatic reasoning given spatially-grounded actions and verbal instructions \cite{buschoff2024visual}. Due to cost and rate limits, we report zero-shot performance with a temperature of zero.

\vspace{6pt}
More details about our models and datasets can be found in the Supplementary Information (\href{https://osf.io/v8ru7/}{\texttt{https://osf.io/v8ru7/}}). Source code is available at 
\href{https://github.com/probcomp/CLIPS.jl}{\texttt{https://github.com/probcomp/CLIPS.jl}}.

\begin{table*}[t]
\centering
\caption{Performance of CLIPS vs. baseline methods, \textmd{measured in terms accuracy (posterior probability of true goal, precision and recall for assistance options), helpfulness (plan length and human cost relative to CLIPS), and human similarity (correlation of goal inferences and assistance options with mean human ratings) Metrics are averaged across the dataset per domain, with standard errors reported in brackets.}}
\begin{tabular}{lrrrrrrr}
\hline
 & \multicolumn{3}{c}{\textbf{Accuracy}} & \multicolumn{2}{c}{\textbf{Helpfulness}} & \multicolumn{2}{c}{\textbf{Human Similarity}} \\
\textbf{Method} & \multicolumn{1}{l}{$P(g_\text{true})$} & \multicolumn{1}{l}{Assist. Prec.} & \multicolumn{1}{l}{Assist. Rec.} & \multicolumn{1}{l}{Rel. Plan Length} & \multicolumn{1}{l}{Rel. Human Cost} & \multicolumn{1}{l}{Goal Cor.} & \multicolumn{1}{l}{Assist. Cor.} \\ \hline
\multicolumn{8}{c}{\textit{Doors, Keys \& Gems}} \\ \hline
Humans & 0.67 (0.04) & 0.83 (0.02) & 0.85 (0.01) & -- & -- & -- & -- \\
CLIPS (Ours) & \textbf{0.74 (0.05)} & \textbf{0.97 (0.03)} & \textbf{0.97 (0.03)} & \textbf{1.00 (0.00)} & \textbf{1.00 (0.00)} & \textbf{0.93 (0.01)} & \textbf{0.96 (0.01)} \\
Lang. Only IP & 0.55 (0.05) & 0.90 (0.06) & 0.83 (0.06) & 1.26 (0.10) & 1.18 (0.07) & 0.74 (0.01) & 0.83 (0.01) \\
Action Only IP & 0.31 (0.04) & 0.43 (0.09) & 0.40 (0.09) & 1.68 (0.15) & 1.46 (0.08) & 0.15 (0.01) & 0.45 (0.01) \\
Literal Efficient & -- & 0.65 (0.08) & 0.54 (0.07) & 1.55 (0.11) & 1.58 (0.10) & -- & 0.47 (0.01) \\
Literal Naive & -- & 0.58 (0.04) & 0.47 (0.02) & 1.54 (0.07) & 1.52 (0.05) & -- & 0.54 (0.01) \\
GPT-4V & 0.29 (0.08) & 0.39 (0.08) & 0.37 (0.08) & -- & -- & \multicolumn{1}{l}{0.10 (0.01)} & 0.11 (0.01) \\ \hline
\multicolumn{8}{c}{\textit{VirtualHome}} \\ \hline
CLIPS (Ours) & \textbf{0.63 (0.04)} & \textbf{0.87 (0.04)} & \textbf{1.00 (0.00)} & \textbf{1.00 (0.00)} & \textbf{1.00 (0.00)} & \textbf{--} & \textbf{--} \\
Lang. Only IP & 0.37 (0.05) & 0.59 (0.05) & 0.96 (0.03) & 1.33 (0.07) & 1.35 (0.08) & -- & -- \\
Action Only IP & 0.25 (0.03) & 0.61 (0.07) & 0.84 (0.07) & 1.30 (0.07) & 1.33 (0.07) & -- & -- \\
Literal Listener & -- & 0.64 (0.06) & 0.70 (0.07) & 1.48 (0.08) & 1.54 (0.09) & -- & -- \\ \hline
\end{tabular}
\label{tab:results}
\end{table*}

\subsection{Human judgments}
\label{sec:human-judgments}

As an additional standard for comparison, we conducted a study with 100 human participants from the US through Prolific (mean age = 39.8, 59 men, 38 women, 2 non-binary), presenting each of them with 15 goal assistance problems from the mDKG domain. In each problem, participants saw the actions taken by the human principal and observed the instruction at the end. Participants were then asked to select the set of gems that they thought were likely to be principal's goal, and then indicate how the robot agent should best assist the principal. This gave 50 goal and assistance ratings for each problem (93\% power for Cohen’s d=0.5 at $\alpha$=0.05). 

For the assistance question, we presented each participant with a set of \emph{assistance options}, corresponding to either picking up keys or unlocking doors. This allowed us to query participants' understanding of how to provide assistance without having them solve the entire problem. To compare these human-provided assistance options with CLIPS and our baselines, we extracted the corresponding actions from the assistive plans generated by each method, and (where applicable) estimated the marginal probability of a particular assistance option occurring in the assistive plan via sampling.

\subsection{Performance metrics}

We evaluated CLIPS and the baseline methods in terms of their goal and assistance accuracy, the helpfulness and efficiency of the generated plans, and their similarity to human judgments. To evaluate accuracy, we calculated the probability $P(g_\text{true})$ assigned to the true goal for the non-literal methods, as well as precision and recall for selecting the optimal assistance options that were consistent with the instruction. To evaluate helpfulness and efficiency, we computed the length of the generated plan and the total action cost incurred by the human principal, relativized to CLIPS.  Finally, to evaluate human similarity in mDKG, we calculated the correlation between each method's outputs with the average ratings provided by our participants. We calculated Pearson's $r$ as well its variability from 1000 bootstrapped samples of the human dataset.

\begin{figure*}[p]
    \begin{subfigure}[t]{0.47\textwidth}
    \includegraphics[width = 8.2cm]{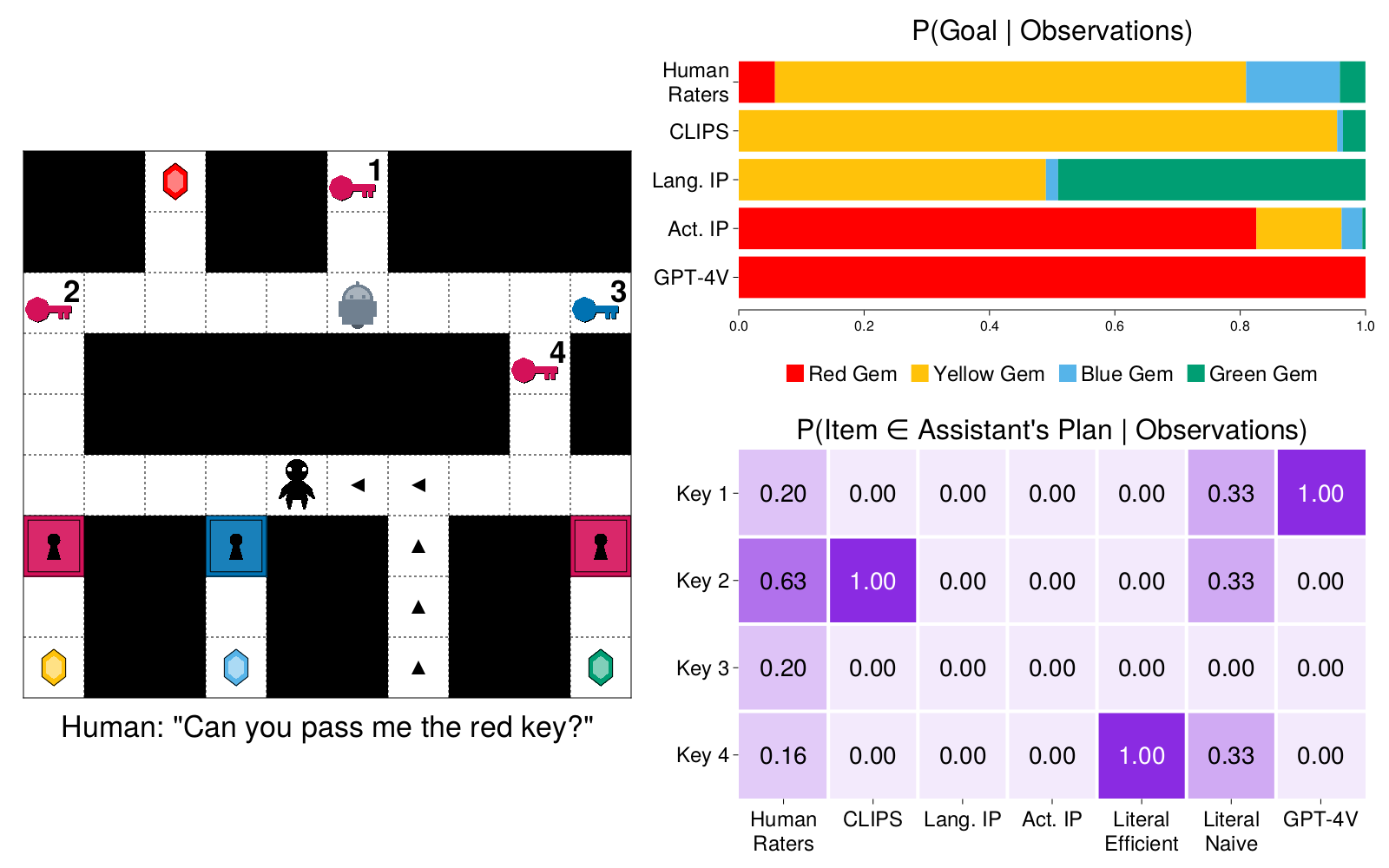}
    \caption{Ambiguous Predicates. \textmd{The human player asks for a red key when three are available. CLIPS resolves this ambiguity, inferring that the human wants \emph{Key 2} to reach the yellow gem.}}
    \label{fig:dkg-ambiguous-predicates}
    \end{subfigure}
    \hspace{0.02\textwidth}
    \begin{subfigure}[t]{0.47\textwidth}
    \includegraphics[width = 8.2cm]{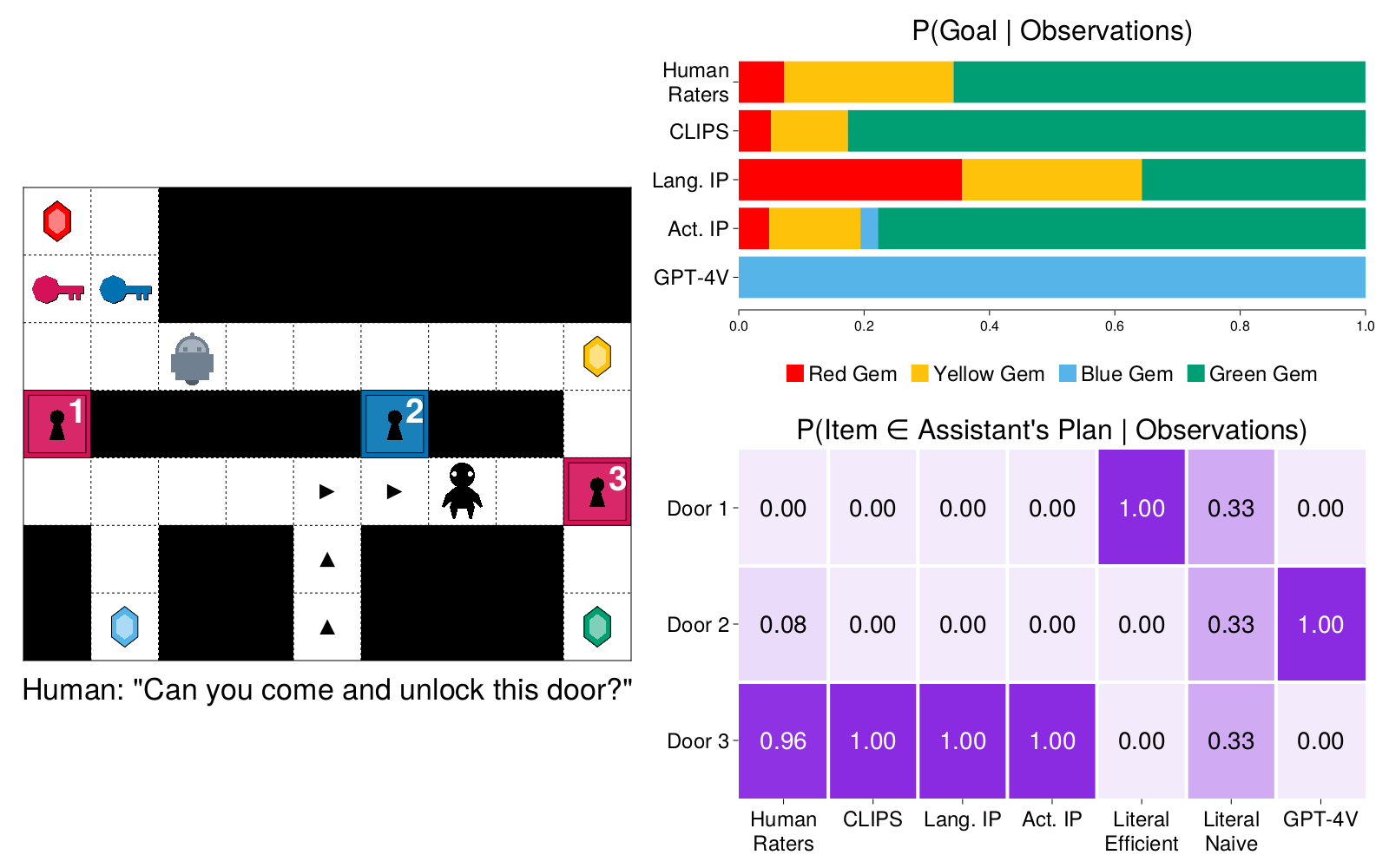}
    \caption{Ambiguous Indexicals. \textmd{The human player refers to a door using the indexical ``this''. CLIPS interprets this pragmatically, inferring that \emph{Door 3} is intended even though \emph{Door 2} is equally close to the human.}}
    \label{fig:dkg-ambiguous-indexicals}
    \end{subfigure}
    
    \begin{subfigure}[t]{0.47\textwidth}
    \includegraphics[width = 8.2cm]{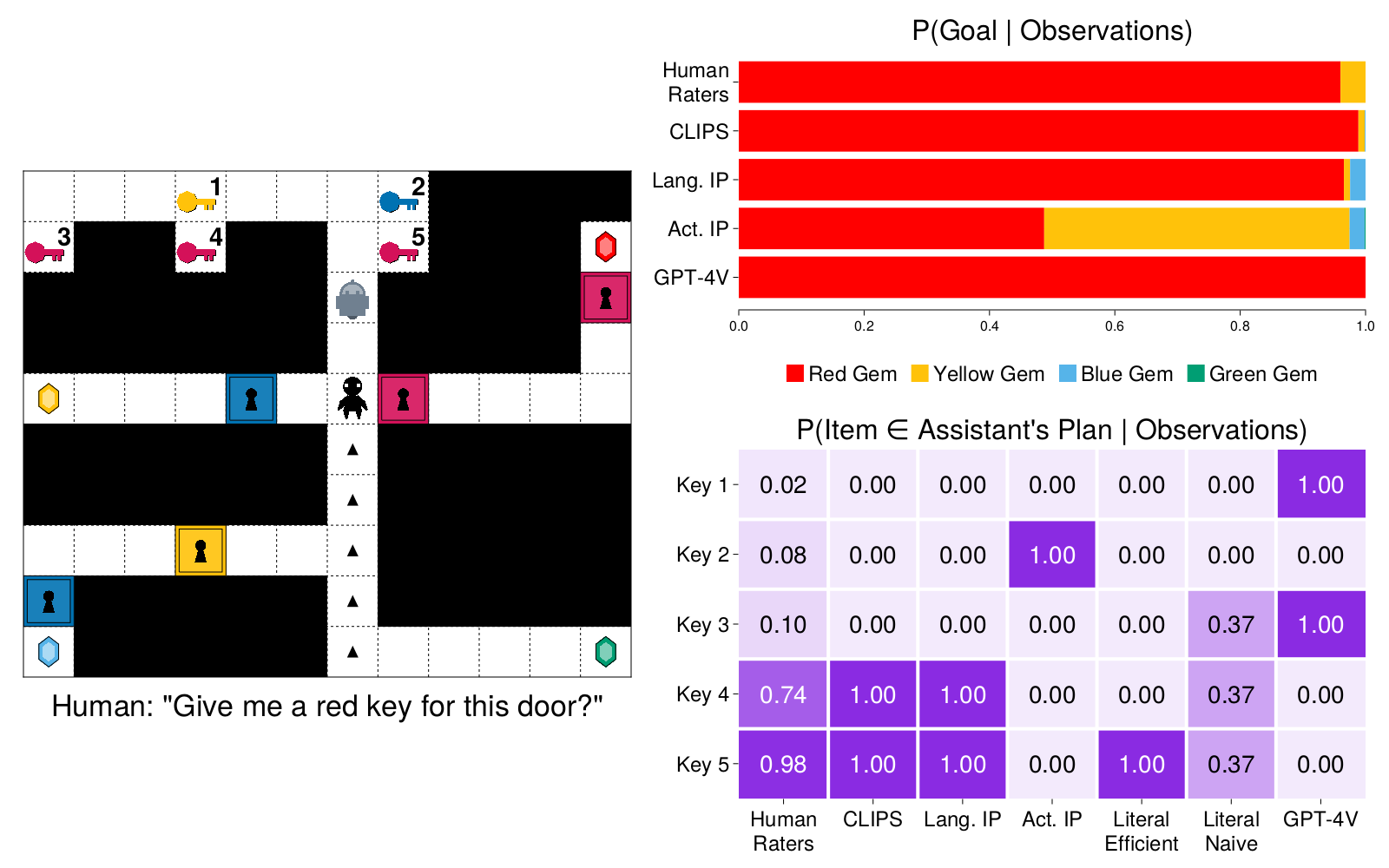}
    \caption{Partial Instructions. \textmd{The human player asks for a single red key, but requires two to reach the red gem. CLIPS infers this and picks up both keys (\emph{Key 4} and \emph{Key 5}), unlike the literal listener baselines.}}
    \label{fig:dkg-partial-instructions}
    \end{subfigure}
    \hspace{0.02\textwidth}
    \begin{subfigure}[t]{0.47\textwidth}
    \includegraphics[width = 8.2cm]{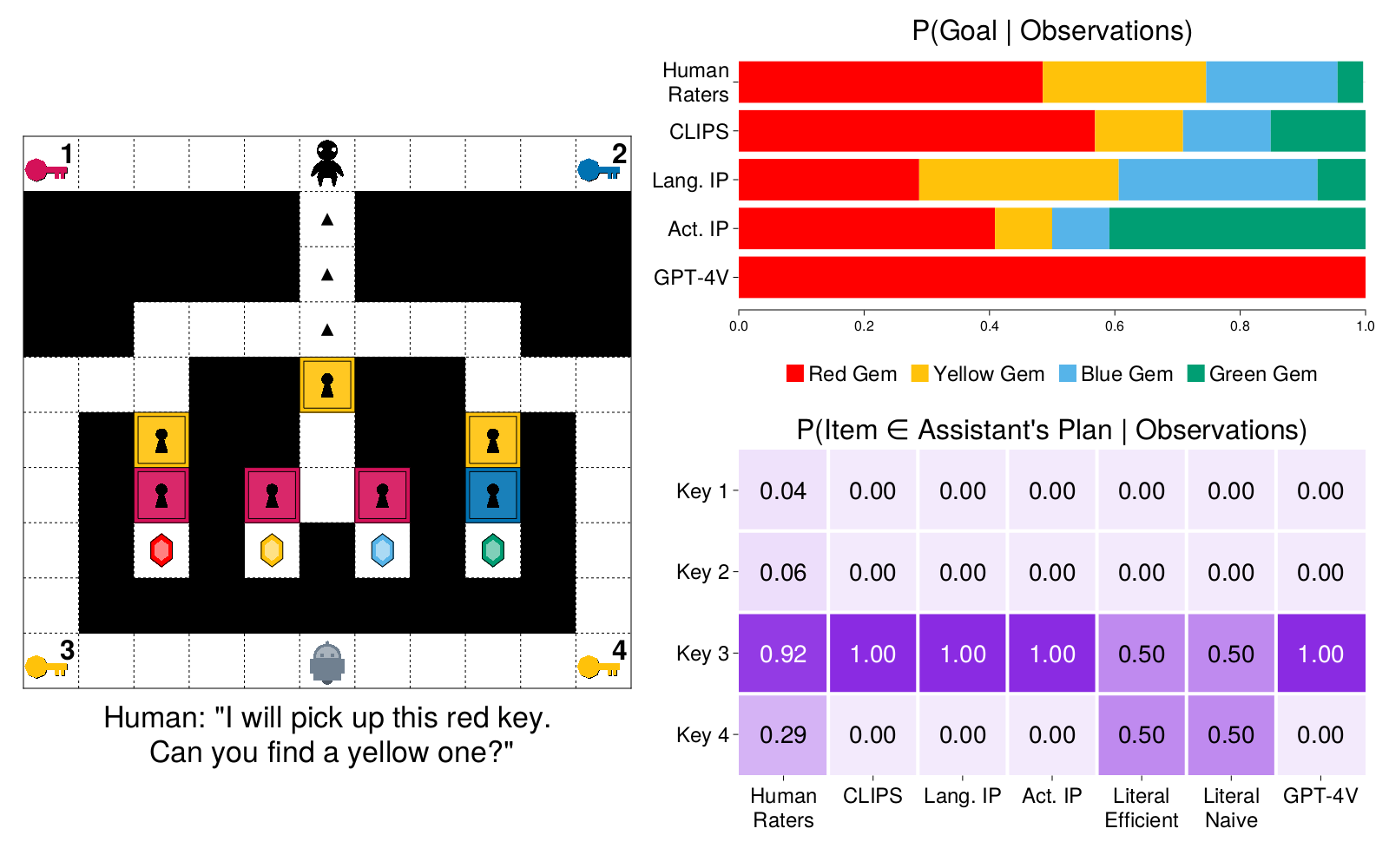}
    \caption{Uncertain Goals. \textmd{The human player asks for a yellow key while going to pick up a red key. CLIPS pragmatically infers their goal to be one of the three gems on the left, and hence picks up the left key (\emph{Key 3}).}}
    \label{fig:dkg-uncertain-goals}
    \end{subfigure}
        
    \begin{subfigure}[t]{0.47\textwidth}
    \includegraphics[width = 8.2cm]{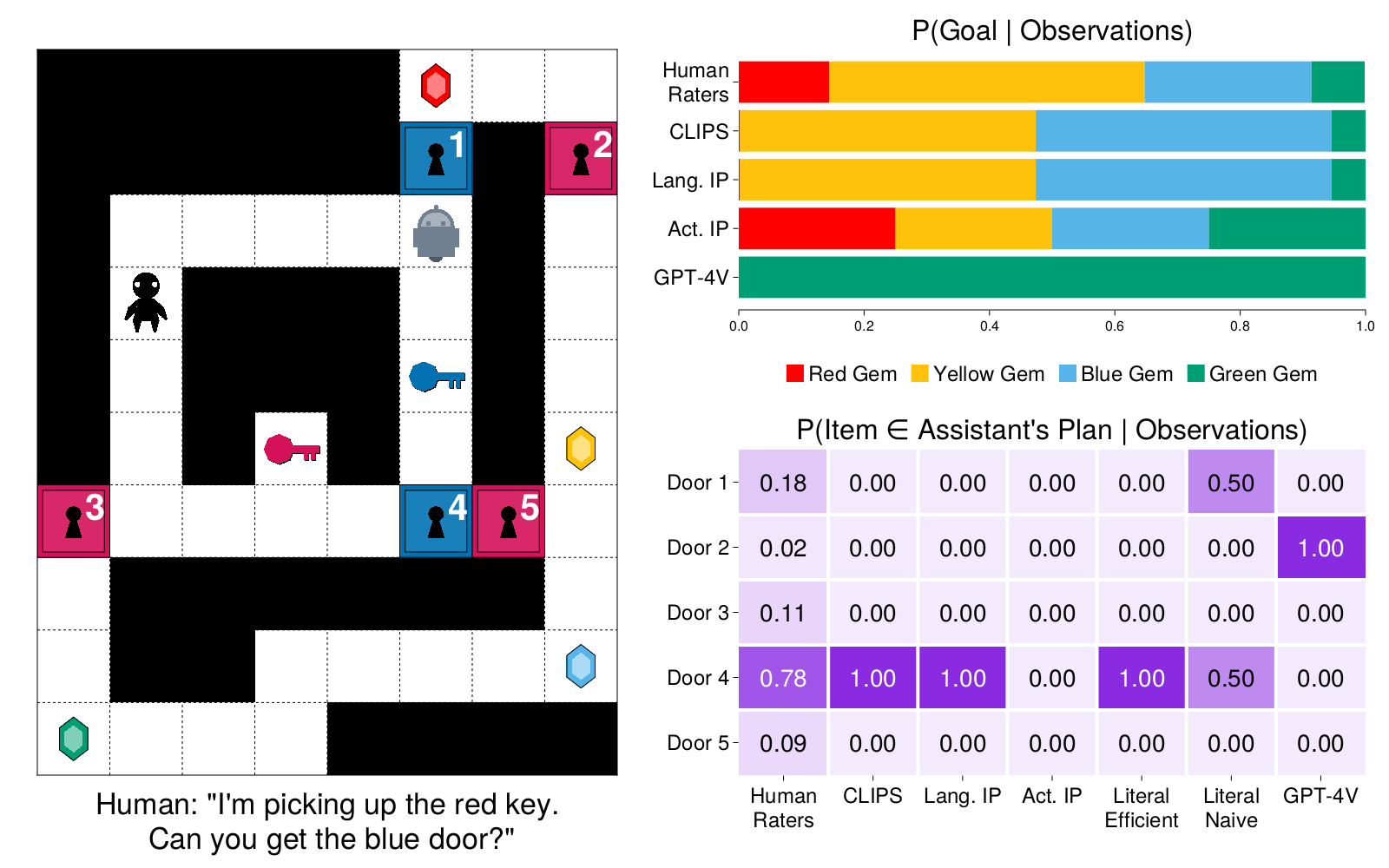}
    \caption{Joint Instructions.\textmd{The human player asks for a blue door to be unlocked while indicating that they will pick up a red key. The CLIPS assistant infers that it should unlock \emph{Door 4} so that the human can unlock \emph{Door 5}.}}
    \label{fig:dkg-joint-instructions}
    \end{subfigure}
    \hspace{0.02\textwidth}
    \begin{subfigure}[t]{0.47\textwidth}
    \includegraphics[width = 8.2cm]{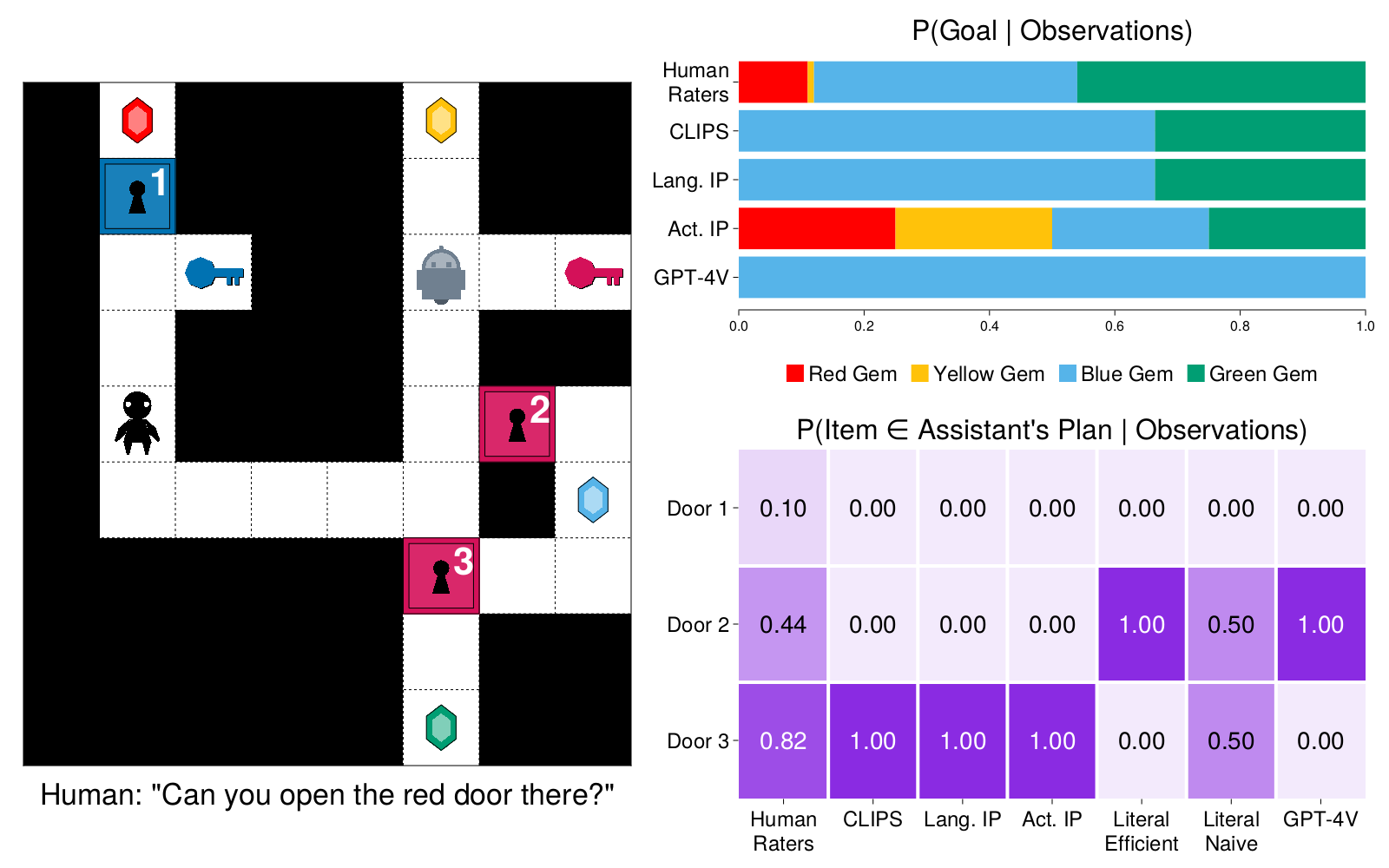}
    \caption{Safe Assistance. \textmd{The human player asks the robot to unlock a red door when there are two such doors. The CLIPS assistant reasons that unlocking \emph{Door 3} is safer as it leads to \emph{both} blue and green gems, unlike \emph{Door 2}.}}
    \label{fig:dkg-safe-assistance}
    \end{subfigure}

    \caption{Goal assistance problems in Doors, Keys \& Gems. \textmd{Each sub-figure contains a visual (\emph{left}), instruction (\emph{bottom left}), goal posteriors produced by each method (\emph{top right}), and the probability of a key or door appearing in the assistive plans generated by each method (\emph{bottom right}). Our pragmatic goal assistance method, CLIPS, best matches the goal inferences and assistance options produced by human raters (averaged across raters). In contrast, language and action-only inverse planning (Lang. IP \& Act. IP) have higher goal uncertainty, the literal baselines fail to resolve instruction ambiguity, and GPT-4V often produces incoherent responses.}}
    \label{fig:dkg-examples}
\end{figure*}

\subsection{Results}

The results of our experiments are presented in Table \ref{tab:results}. In terms of accuracy, CLIPS assigned higher probability to the true goal than the baselines (note that 100\% is not possible since goals cannot always be distinguished, as in Fig. \ref{fig:dkg-uncertain-goals}), while achieving close to perfect precision and recall in selecting the optimal assistance options. Indeed, CLIPS outperformed the average human in this regard. In contrast, unimodal inverse planning was much less accurate at inferring the true goal, and all baselines selected the correct assistance options at significantly lower rates. This affected the efficiency of assistance: CLIPS was 1.26 to 1.68 times faster in achieving the human's goal than the baselines, and produced plans that were 1.18 to 1.58 times less costly for the human.

For human similarity in the mDKG domain, we found that both the goal inferences and the assistance options produced by CLIPS correlated highly with human ratings, achieving Pearson's $r$ of 0.93 (95\% CI: 0.91--0.94) and 0.96 (95\% CI: 0.95--0.96) respectively. In contrast, unimodal inverse planning produced goal inferences that were highly dissimilar from humans. Correlation with human-selected assistance options was also poor. These results demonstrate the importance of accounting for pragmatic context. Out of all methods, GPT-4V performed the worst, possibly because of the spatial reasoning and multi-step planning required for coherent goal inference in mDKG. Among participants themselves, we found that the median Pearson’s $r$ of each participant’s ratings with the mean rating was 0.85 (IQR: 0.71--0.94) for goal inferences and 0.87 (IQR: 0.65--0.95) for assistance options, indicating that our dataset of human judgments is a reliable measure of average human performance.

In Figure \ref{fig:dkg-examples}, we compare the results of CLIPS against human judgements and baselines on six illustrative goal assistance problems, providing qualitative observations in the captions. The results show how CLIPS closely mirrors human judgements for both goal inference and assistance, resolving ambiguity in predicates (Fig. \ref{fig:dkg-ambiguous-predicates}) and indexicals (Fig. \ref{fig:dkg-ambiguous-indexicals}) while successfully completing partial instructions (Fig. \ref{fig:dkg-partial-instructions}) and interpreting joint instructions (Fig. \ref{fig:dkg-joint-instructions}). In comparison, the unimodal and literal baselines make less confident inferences or fail to assist appropriately, while GPT-4V provides incoherent answers. We also find that both humans and CLIPS are able to assist appropriately when there is significant goal uncertainty (Fig. \ref{fig:dkg-uncertain-goals}), and even when the optimal assistive plans for each goal make diverging recommendations (Fig. \ref{fig:dkg-safe-assistance}), illustrating the importance of uncertainty-aware assistance.

\section{Related Work}

\paragraph{Theory of Mind as Bayesian Inverse Planning}
Bayesian Theory of Mind (BToM) posits that humans reason about the actions and mental states of others through (approximately) Bayesian inference \cite{baker2011bayesian,baker2017rational}.
In particular, Bayesian inverse planning can model how humans infer the goals of others by assuming that agents act rationally to achieve their goals \cite{gergely2003teleological,baker2009action,shum2019theory, jara2019naive}. To efficiently solve these inference problems, prior work uses sequential Monte Carlo methods \cite{del2006sequential,lew2023smcp3} to perform online goal inference over model-based planning agents \cite{zhi2020online, alanqary2021modeling}. CLIPS builds upon this paradigm, extending recent work by \citet{ying2023inferring} on multi-agent inverse planning.

\paragraph{Rational Speech Acts}
Rational Speech Act (RSA) theory frames communication as a recursive process, where speakers make pragmatic utterances that optimize relevance to listeners, and listeners interpret speaker utterances in light of the pragmatic goals that underlie those utterances \cite{goodman2016pragmatic,fried2018unified,sumers2023reconciling}. CLIPS instantiates the RSA framework in the context of cooperative planning.

\paragraph{Multimodal Goal Inference and Reward Learning.} Goal inference can be framed as online inverse reinforcement learning (IRL), where the aim is to to learn a reward or cost function explaining the agent's behavior in a single episode \cite{jara2019theory}. CLIPS can thus be viewed as a language-informed IRL method, though IRL is primarily applied offline given a dataset of expert demonstrations \cite{williams2018learning,tung2018reward,fu2019language}. Particularly relevant is \emph{reward-rational implicit choice} \cite{jeon2020reward}, a framework for multimodal Bayesian reward learning from heterogeneous human feedback. CLIPS can be seen as an application of this framework for specific modalities, but extended to the cooperative setting.

\paragraph{Decentralized Cooperation and Joint Intentionality}
When multiple agents cooperate on a task, they usually need to track each other's mental states to make decisions. Common approaches to this involve recursive reasoning over each other's goals and plans, which can quickly grow intractable \cite{doshi2009monte,camerer2004cognitive}. To address this challenge, computational cognitive scientists have developed models of \emph{joint intentionality} \cite{tomasello2007shared}, where cooperating agents conceive of themselves as a \emph{group agent} with a common goal in mind \cite{stacy2021modeling,tang2022exploring}. This approach allows groups of agents to converge on shared goals \cite{tang2020bootstrapping} and efficient task decompositions \cite{wu2021too} in a decentralized manner with limited recursion. CLIPS uses this insight to model joint planning and pragmatic instruction generation in humans.

\paragraph{Value Alignment and Assistance Games.} CLIPS is a solution strategy for language-augmented goal assistance games, an extension of the assistance game formalism for human-AI value alignment \cite{hadfield2016cooperative,russell2021human}. In contrast to prior work, our framework leverages joint intentionality when modeling human principals, thereby requiring less recursion than iterated best-response \cite{hadfield2016cooperative}, while avoiding the intractability of equilibrium solutions \cite{fisac2017pragmatic,milli2020literal}.

\paragraph{Instruction Following with Language Models.} CLIPS is a form of grounded instruction following from natural language \cite{tellex2011understanding,squire2015grounding,misra2016tell}, 
using LLMs to score the likelihood of an utterance given a grounded command. While many recent studies have employed LLMs for translating language into actions \cite{ahn2022can,singh2023progprompt} or task specifications \cite{kwon2023reward,yu2023language,liu2022lang2ltl}, we opt for an explicitly Bayesian approach, allowing our method to integrate information from both actions and instructions in a principled, reliable, and modular manner.

\section{Discussion and Future Work}

In this paper, we introduced cooperative language-guided inverse plan search (CLIPS) as a Bayesian architecture for pragmatic instruction following and goal assistance. By using a structured cognitive model of how speakers produce both actions and utterances given their goals, CLIPS is able to integrate information from observed actions and ambiguous instruction, inferring a distribution over speaker's goals and intentions. Using this distribution, a CLIPS assistant is then able to assist the speaker in achieving their goal through expected cost minimization. Through our experiments, we show that the proposed architecture produces human-like outputs on goal inference and assistance tasks. Compared with the baselines, CLIPS also makes more accurate goal inferences, and chooses actions that result in more efficient plans.

While CLIPS demonstrates compelling theoretical properties and strong empirical performance, a number of challenges remain before it can be applied to scenarios with a larger number of goals and assistance options. As noted in Section \ref{sec:goal-inference}, the Bayesian inference strategy we adopt is fully enumerative, but this approach breaks down once the space of possible goals, plans, and commands grows sufficiently large. Recent advances in probabilistic programming could overcome these bottlenecks. For example, more sophisticated sequential Monte Carlo strategies could be used to flexibly propose and update hypotheses, focusing computation on only the most likely sets of latent variables \cite{zhi2020online,lew2023smcp3}. These strategies can be applied to perform \emph{constrained decoding} from large language models \cite{lew2023sequential,willard2023efficient}, thereby allowing LLMs to be used as sound proposal distributions over a grammar for commands. This would benefit from LLMs' performance at few-shot semantic parsing and translation \cite{shin2021constrained}, while preserving the Bayesian nature of our framework.

The assistance policy we introduced in Algorithm \ref{alg:clips-assistance} could also be extended in a number of ways. In many assistive settings, uncertainty is sufficiently high that it makes sense for the assistant to take \emph{information gathering actions}. Indeed, this is what people typically do when we are confused: we ask questions. To enable this ability, assistants should perform \emph{belief-space planning} over the set of possible goal beliefs \cite{sunberg2018online,macindoe2012pomcop}, thinking ahead about which actions reveal more information about the principal's goals. Such planning would alleviate issues that can arise due to the $Q_\text{MDP}$ approximation \cite{littman1995learning,hauskrecht2000value} used by Algorithm \ref{alg:clips-assistance}. In the longer run, assistants could be augmented with natural language outputs, enabling them to ask clarifying questions by planning ahead in a white-box, interpretable manner. If successful, this would constitute a considerable step towards trustworthy assistive AI that effectively collaborates and communicates with humans.

\newcommand\extrafootertext[1]{%
    \bgroup
    \renewcommand\thefootnote{\fnsymbol{footnote}}%
    \renewcommand\thempfootnote{\fnsymbol{mpfootnote}}%
    \footnotetext[0]{#1}%
    \egroup
}

\extrafootertext{\textbf{Errata:} Along with typographical fixes, this version corrects minor errors in the GPT-4V assistance results that are present in the version of this paper published in the ACM Digital Library.}

\begin{acks}
This work was funded in part by the DARPA Machine Common Sense, AFOSR, and ONR Science of AI programs, along with the MIT-IBM Watson AI Lab, and gifts from Reid Hoffman and the Siegel Family Foundation. Tan Zhi-Xuan is supported by an Open Philanthropy AI Fellowship.
\end{acks}

\bibliographystyle{ACM-Reference-Format} 
\bibliography{sources}


\pagebreak
\onecolumn
\appendix

\setcounter{figure}{0}
\setcounter{table}{0}
\renewcommand\thefigure{\thesection\arabic{figure}}
\renewcommand\thetable{\thesection\arabic{table}}

\section{Goal Assistance Problems}

Table \ref{tab:dkg-problems} and Table \ref{tab:vh-problems} describe the scenarios and conditions for our goal assistance problems in multi-agent Doors, Keys \& Gems (mDKG) and VirtualHome (VH) respectively. VH scenarios were encoded in PDDL \cite{mcdermott1998pddl} for use with our inverse planning system, but they can be easily be translated to VirtualHome's evolving graph simulator or Unity simulator \cite{puig2018virtualhome}.

\begin{table}[h]
\centering
\caption{Goal assistance problems in Doors, Keys \& Gems. \textmd{30 problems were divided into those that required assistance with either picking up keys or unlocking doors. Actions and utterances were varied according to the following factors.}}
\begin{tabular}{@{}llp{12cm}@{}}
\toprule
\textbf{Factor} & \textbf{Condition} & \textbf{Description} \\ \midrule
\multirow{2}{*}{\begin{tabular}[c]{@{}l@{}}Instruction\\ Uncertainty\end{tabular}} & Ambiguous & Ambiguous instruction that can be associated with many possible items. (e.g. \emph{``Can you get me the red key?''} when multiple red keys are available.)\vspace{3pt} \\
 & Partial & The principal only requests a subset of the items they require to achieve their goal. (e.g. \emph{``Can you get me the red key?''} when both red and blue keys to reach its goal.)\vspace{3pt} \\ \midrule
\multirow{2}{*}{\begin{tabular}[c]{@{}l@{}}Agent\\ Involvement\end{tabular}} & Assistant Actions & The instruction involves only actions that the assistant should take. (e.g. \emph{``Can you pick up that key?''}, \emph{``Please unlock the red doors.''})\vspace{3pt}\\
 & Joint Actions & The instructions contain information about both agents' actions. (e.g. \emph{``I'm getting the red key, can you get the blue key?''})\vspace{3pt} \\ \midrule
\multirow{2}{*}{\begin{tabular}[c]{@{}l@{}}Goal\\ Uncertainty\end{tabular}} & Low Uncertainty & The actions and instructions combined are largely sufficient to uniquely determine the goal. (e.g. In Figure \ref{fig:dkg-ambiguous-indexicals}, only the red gem is likely.) \vspace{3pt} \\
 & High Uncertainty & The principal's goal remains ambiguous even after observing both their actions and instructions. (e.g. In Figure \ref{fig:dkg-uncertain-goals}, the three gems on the left are plausible goals.) \vspace{3pt} \\ \bottomrule
\end{tabular}
\label{tab:dkg-problems}
\end{table}

\begin{table}[h]
\centering
\caption{Goal assistance problems in VirtualHome. \textmd{25 problems were divided into \textbf{(a)} three scenarios, which correspond to different initial states of the environment, and had different sets of possible goals. Actions and utterances \textbf{(b)} were varied across several conditions.}}
\label{tab:vh-problems}
\begin{subtable}[b]{\textwidth}
\centering
\begin{tabular}{@{}lll@{}}
\toprule
\textbf{Scenario} & \textbf{Goal Category ($k$)} & \textbf{Description} \\ \midrule
Food Preparation & Salads (3) & Place a chef's knife \& ingredients (onion, cucumber/potato \& tomato/chicken) on the table. \\
\textit{(6 goals)} & Stews (3) & Place an onion, carrot, a main (salmon, chicken, or potato) \& wine on the table. \\ \midrule
Table Setting & Plates (4) & Place $n$ plates on the table, where $n \in [1, 4]$. \\
\textit{(12 goals)} & Plates \& Cutlery (4) & Place $n$ plates, $n$ forks \& $n$ knives on the table, where $n \in [1, 4]$. \\
 & Plates \& Bowls (4) & Place $n$ plates, $n$ forks, $n$ knives \& $n$ bowls on the table, where $n \in [1, 4]$. \\ \midrule
Drink Pairing & Wine (3) & Place wine and $n$ wineglasses on the table, $n \in [1, 3]$. \\
\textit{(12 goals)} & Wine Pairing (3) & Place wine, cheese, $n$ wineglasses \& $n$ forks on the table, $n \in [1, 3]$. \\
 & Juice (3) & Place juice \& $n$ waterglasses on the table, $n \in [1, 3]$. \\
 & Juice Pairing (3) & Place juice, $n$ waterglasses, $n$ cupcakes \& $n$ forks on the table, $n \in [1, 3]$. \\ \bottomrule
\end{tabular}
\vspace{1pt}
\caption{\textmd{Scenarios and goals.}}
\end{subtable}
\begin{subtable}[b]{\textwidth}
\centering
\begin{tabular}{@{}llp{12.4cm}@{}}
\toprule
\textbf{Factor} & \textbf{Condition} & \textbf{Description} \\ \midrule
\multirow{2}{*}{\begin{tabular}[c]{@{}l@{}}Instruction\\Uncertainty\end{tabular}} & Ambiguous Number & Instruction which mentions an ambiguous number of items to be collected. (e.g. \emph{``Can you help with these plates?''}) \vspace{3pt} \\
 & Ambiguous Type & Instruction which is ambiguous about the type of item to be collected. (e.g. \emph{``Can you get two glasses?''}, \emph{``And I need the veggies for the stew.''}) \vspace{3pt} \\ \midrule
\multirow{2}{*}{\begin{tabular}[c]{@{}l@{}}Utterance\\Count\end{tabular}} & Single & The principal gives a single instruction to the assistant. (e.g. \emph{``Go get the forks and knives.''}, \emph{``Can you get some wine?''}) \vspace{3pt} \\
 & Multiple & The principal gives an instruction to the assistant, takes a few more actions, then gives another instruction. (e.g. \emph{``Could you find two glasses?''} followed by \emph{``Get some forks as well.''}) \vspace{3pt} \\ \bottomrule
\end{tabular}
\vspace{1pt}
\caption{\textmd{Factors and conditions.}}
\end{subtable}
\end{table}

\section{Model Configuration}

In this section, we provide further details about how CLIPS and each of our baselines was configured for each domain.

\subsection{CLIPS \& Unimodal Inverse Planning}

We describe the configuration for CLIPS and unimodal inverse planning together, since the only difference between the former and the latter is whether both action likelihoods $P(a^{_h}_t | s_t, \pi)$ and utterance likelihoods $P(u_t | s_t, \pi)$ are incorporated into the weight $w$ associated with each hypothesized goal $g$, cost function $C$ and policy $\pi$ in Algorithm \ref{alg:clips-inference}.

\subsubsection*{\textbf{Goal and Cost Priors}}

In both domains, we assumed a uniform goal prior $P(g)$ over the set of possible goals (4 in mDKG, 6--12 in VH). In mDKG, we considered four possible cost functions $C \in \mathsf{C}$, and placed a uniform prior $P(C)$ over such cost functions. This made CLIPS more robust to the possibility that the human would prefer to do less (or more) work, asking the robot to pick up keys or unlock doors that would be sub-optimal if action costs were equal between both agents. The costs for each action were as follows:
\begin{itemize}
    \item Human pick-up: \textbf{5}. Robot pick-up: \textbf{1}. Wait cost: \textbf{0.6}. All other action costs: \textbf{1}.
    \item Human pick-up: \textbf{5}. Robot pick-up: \textbf{1}. Wait cost: \textbf{0.6}. Other human costs: \textbf{2}. Other robot costs: \textbf{1}.
    \item Human pick-up: \textbf{5}. Robot pick-up: \textbf{1}. Robot unlock: \textbf{5}. Wait cost: \textbf{0.6}. All other action costs: \textbf{1}.
    \item Human pick-up: \textbf{5}. Robot pick-up: \textbf{1}. Robot unlock: \textbf{5}. Wait cost: \textbf{0.6}. Other human costs: \textbf{2}. Other robot costs: \textbf{1}.
\end{itemize}

In VH, we avoided the need the perform online inference over multiple cost functions by using a slightly more flexible utterance model that allowed for the possibility that the principal might ask the assistant to carry out sub-optimal actions (e.g. asking the assistant to acquire \emph{all} remaining items, even though the optimal joint plan would have both agents share the work more equally). This allowed us to use a \emph{single} cost function that assigned unit cost to all actions for both agents. Future work should investigate how more flexible models for multi-agent task division (e.g. \cite{wu2021too}) could be integrated with CLIPS, along with offline learning of cost function priors.

\subsubsection*{\textbf{Policy Configuration}}

In Bayesian inverse planning, an algorithm is required to compute the joint policy $\pi$ for each possible goal. While it is possible to do this using a domain-general planning algorithm and search heuristic, we configured our planners in a domain-specific manner to achieve greater speed and reliability.

\paragraph{Doors, Keys \& Gems} In mDKG, we used the Real-Time Adaptive A* (RTAA*) variant \cite{koenig2006real} of real-time heuristic search \cite{korf1990real} to incrementally compute the value function $V(s)$ for the policy $\pi$. Given $V(s)$, one can then compute $Q(s, a) = C(s, a) + \sum_{s'} P(s'|s, a) V(s')$. Each time \textsc{rths-policy-update} is called, we perform A* search up to a search budget of $2^{18}$ nodes from each state $s_n$ that neighbors the current state $s$, using the current estimate $\hat V$ of the value function as the search heuristic. Once A* search terminates at state $s_f$ (either by reaching the goal or using up the budget), we update the value function for \emph{all} states $s_i$ in the interior of the search tree as follows:
\[\hat V(s_i) \gets \hat V(s_f) - C(s_n, s_f) + C(s_n, s_i)\]
where $C(s_n, s_i)$ is the cost of getting from $s_n$ to $s_i$ by traversing the search tree. As shown in \cite{koenig2006real}, as long as $\hat V$ is initialized to a \emph{consistent} (under-)estimate of the true value function, subsequent $\hat V$ values will remain consistent. We ensured this by (lazily) initializing $\hat V$ with a minimum spanning tree heuristic \cite{held1971traveling} over the graph of keys and doors required to reach the goal, using a preprocessing step used in inventory-aware pathfinding \cite{aversa2016pruning} to extract necessary keys and doors. To further accelerate planning, we pruned actions that were irrelevant to the goal (e.g. picking up unnecessary keys), and implemented cost-minimal path reuse from the tree-adaptive variant of RTAA* \cite{hernandez2015reusing}, allowing us to terminate A* search whenever an existing path to the goal was found. Collectively, these algorithmic choices allowed us to estimate the optimal value function $V^*$ in close to real-time for most timesteps, despite the difficulty of optimal multi-agent planning.

\paragraph{VirtualHome} In VH, we also experimented with using RTAA* to compute the policy $\pi$. However, this unpredictably increased the latency of inverse planning when the search budget was too high, and led to unstable $Q$-value estimates when the search budget was too low. Instead we found that it was sufficient to compute an analytic lower bound $\hat V$ to the optimal value function $V^*$ by counting the number of remaining items and movements required to achieve the goal. Using $\hat V$ was both faster (no search was required), and produced more consistent estimates of the relative $Q$-values $Q(s, a) - V(s)$ (also known as the \emph{advantage function}) for each state $s$ and goal $g$, resulting in more stable action likelihoods $P(a | s, \pi)$ and goal inferences. However, this required hand-deriving a lower bound on $V^*$. In the future, we hope to explore the use of domain-general lower bounds on $V^*$ using admissible planning heuristics \cite{haslum2005new,helmert2009landmarks}, and to implement more consistent value function updates based on the LSS-LRTA* variant of real-time heuristic search \cite{koenig2009comparing}, allowing us to stably estimate $V^*$ with a low search budget.

\paragraph{Rationality Modeling} As in hierarchical Bayesian approaches to inverse reinforcement learning \cite{choi2014hierarchical} and theory-of-mind \cite{zhi2022solving}, we placed a Gamma prior over the rationality parameter $\beta$ for the Boltzmann policy, discretized across 33 values from $2^{-3}$ to $2^{5}$ by incrementing the exponent with a step size of $0.25$. In mDKG, we used a $\text{Gamma}(0.5, 1)$ prior, while in VH, we used a $\text{Gamma}(1, 2)$ to match the less optimal actions taken in that domain. For efficient inference, we used Rao-Blackwellization \cite{murphy2001rao}, marginalizing over all values of $\beta$ when computing the action likelihood $P(a^{_h} | s_t, \pi)$, then analytically updating the local posterior $P(\beta | a^{_h}_{1:t}, s_{1:t}, \pi, g)$.

\subsubsection*{\textbf{Utterance Model}}

In the CLIPS utterance model, salient actions $\alpha_{1:m}$ are first extracted from the joint policy $\pi$, before being sub-sampled into a command $c$, which is then translated into an utterance $u$. In our implementation, we enumerated over commands to reduce variance. This section describes how we extracted salient actions, the structure of commands, and the command enumeration strategy for each domain.

\paragraph{Extracting Salient Actions} Following work in linguistics \cite{giora2004graded,kecskes2004role} and RSA theory \cite{goodman2016pragmatic}, we assumed that some types of actions were more likely to be communicated because of their \emph{salience} to both the speaker and the listener. To model this, after rolling out the joint policy $\pi$ for up to $H$ steps, we filtered out all non-salient actions, leaving only the salient actions  $\alpha_{1:m}$ to be communicated. In VH, to account for the possibility of the principal delegating \emph{all} remaining items to the assistant (instead of equally sharing the load), we enforced that the principal could not collect new items when rolling out the policy.

As in \cite{ying2023inferring}, we manually defined action salience: In mDKG, \texttt{pickup}, \texttt{unlock}, and \texttt{handover} actions and \texttt{iscolor} predicates were salient, while in VH, \texttt{grab} actions and \texttt{itemtype} predicates were salient. In general, however, action salience could be learned, either from a corpus, or by offline estimation of which actions tend be more \emph{unique} and hence \emph{informative} for a particular goal and plan.

\paragraph{Command Enumeration} Formally, we define a command $c := ([a, ...], \{p, ...\}, \{(v, \tau), ...\})$ as a sequence of (possibly lifted) actions $[a, ...]$, a set of associated predicates $\{p, ...\}$ that qualify an action's arguments, and a set $\{(v, \tau), ...\}$ of free variables $v$ with associated types $\tau$ which appear as an argument in at least one action in  $[a, ...]$. For example, the command {\small\texttt{(handover robot human ?key1) where (iscolor ?key1 blue)}} has one partially lifted action $a =$ {\small \texttt{(handover robot human ?key1)}}, along with a predicate $p =$ {\small\texttt{(iscolor ?key1 blue)}} that describes the variable $v =$ {\small\texttt{?key1}} as blue, where the type of {\small\texttt{?key1}} is $\tau =$ {\small\texttt{key}}.

To generate commands from a set of a salient actions $\alpha_{1:m}$ and associated predicates $p_{1:m}$, we enumerated over all subsets of salient action-predicate pairs of up to size $K = 3$ in mDKG and $K = 4$ in VH, before pruning away combinations of actions that were less likely. In mDKG, we did this by excluding commands with more than two distinct action types, commands where \emph{all} actions were taken by the speaker/principal, and also commands which involved chains of dependent actions (e.g. picking up a key, followed by handing over the same key). In VH, we excluded all commands that involved \emph{any} action by the speaker (since our VH dataset did not contain joint instructions). Following enumeration of these (ground) commands, we then lifted the commands, replacing concrete object names with corresponding typed variables. We also replaced all agent identifiers such as \texttt{human} and \texttt{robot} with corresponding indexicals (\texttt{me} and \texttt{you}). Duplicate lifted commands were then removed.

These choices were made partly to limit the number of commands enumerated. Future extensions to CLIPS that make use of constrained LLM decoding \cite{lew2023sequential,shin2021constrained} could avoid the limitations involved in command enumeration, directly translating an utterance $u$ into a much larger space of possible commands (e.g. a context-free grammar over commands).

\paragraph{Language Models} In mDKG, we used OpenAI's 6.7B \texttt{curie} model \cite{brown2020language} as our LLM\footnote{Unfortunately, as of 4 January 2024, OpenAI has shutdown access to \texttt{curie} and the original GPT-3 family of models, making it no longer possible to reproduce our results exactly. Nonetheless, we expect similar results when using \texttt{davinci-002} as a replacement model.} , while in VH, we used their more recently released \texttt{davinci-002} model. Both of these were base models, providing more diffuse probabilities that avoid issues with over-confidence and mode collapse that can arise in LLMs finetuned on instructions or human preferences \cite{hopkins2023can,kirk2023understanding}.

LLMs were instantiated as generative functions in the Gen probabilistic programming system \cite{cusumano2019gen} using the GenGPT3.jl library \cite{zhi2022gengpt3}, allowing us to easily score the probability $P(u | c, \mathcal{E})$ of an utterance $u$ given a command $c$ and few-shot examples $\mathcal{E}$ provided in the prompt. Each utterance was scored against all enumerated commands in parallel via a batched API request. In mDKG, we provided 18 examples of commands being translated to utterances, while in VH, we provided 12 examples. Examples can be found in the source code.

\subsubsection*{\textbf{Assistance Policy}}

When running the $Q_\text{MDP}$ assistance policy shown in Algorithm \ref{alg:clips-assistance}, we filtered out goal hypotheses with a weight $w^i$ of less than $0.02$, so as to avoid the cost of updating the policies $\pi^i$ for unlikely goals. In mDKG, we updated the policies using RTAA* with a node budget $2^16$ and a time limit of $10\text{s}$. In VH, we updated $Q$-values using the analytic lower bound $\hat V$ mentioned earlier.

We evaluated the assistance policy against a simulated human principal, which was designed to (i) follow a ``ground truth'' plan whenever possible, (ii) switch to following an approximately optimal joint policy when the following the original plan was not possible, (iii) fallback to a single-agent policy upon detection that the assistant was not following the joint policy to the true goal (i.e. not being helpful). To detect this, the principal compared how likely the assistant was to be following a joint policy to the true goal (Boltzmann, with $\beta = 1$) vs. a random baseline, switching to the fallback policy once it was 10 times more likely for the assistant to be acting at random.

\subsection{LLM-Based Literal Listener}

Our literal listener baselines consisted of two components: literal command inference, and literal assistance. We describe each below.

\subsubsection*{\textbf{Literal Command Inference}}

Command inference was implemented similarly to inference over the CLIPS utterance model. We first enumerated over a set of possible commands in the current state $s$, then used a LLM to score the probability $P(u | c, s)$ of the observed utterance $u$ for each command $c$. Normalizing these probabilities produced a posterior over commands $P(c | u, s)$.

\paragraph{Utterances} In mDKG, we made command inference easier by replacing joint instructions (e.g. \emph{``I'm getting the red key, can you get the blue key?''}) with assistant-directed instructions (e.g. \emph{``Can you get the blue key?''}). In VH, we handled multiple utterances $u_{t_1}$, $u_{t_2}$ by merging the inferred commands $c = c_{t_1} \cup c_{t_2}$. This created a product distribution $P(c | u_{t_1}, u_{t_2}, s_{t_1}, s_{t_2}) = P(c_{t_1} | u_{t_1}, s_{t_1}) P(c_{t_2} | u_{t_2}, s_{t_2})$ over merged commands.

\paragraph{Command Enumeration} To enumerate commands, we first enumerated all salient actions that were reachable from the current state $s$. We then enumerated all subsets the set of salient actions of up to size $K = 3$ in mDKG an $K = 4$, pruning away certain combinations as in the CLIPS utterance model. Commands were then lifted, and duplicate commands were removed.

\paragraph{Language Models} We used \texttt{text-davinci-002} for mDKG. Given the larger space of commands for the literal listener, we found that using a RLHF finetuned model such as \texttt{text-davinci-002} ensured that more posterior probability mass concentrated on the intuitively correct set of commands. To simulate this effect in our VH experiments after \texttt{text-davinci-002} access was shutdown by OpenAI, we used \texttt{davinci-002} with a  temperature of 0.5. 12 few-shot examples were used for both mDKG and VH.

\subsubsection*{\textbf{Literal Assistance}}

Literal assistance involved several steps: (i) Systematically sampling a command $c$ from the inferred distribution $P(c | u, s)$; (ii) Converting the command $c$ to an equivalent goal formula $g_c$; (iii) Generating an assistive plan that first achieved $g_c$, then the true goal $g$.

\paragraph{Sampling Commands} As described in the main text, we estimated the \emph{mean} performance of literal assistance by drawing $m = 10$ samples from the distribution $P(c | u, s)$ and averaging our metrics across these samples. To do this in a low-variance manner, we first sorted the commands by their inferred probabilities. We then used systematic sampling \cite{yates1948systematic}, selecting a command $c_i$ as long the random variable $x_i$ fell within its corresponding probability bin, where $x_i = (i-1)/m + x$ for $i \in [1, m]$, and $x \sim U(0, 1/m)$.

\paragraph{Mapping Commands to Goals} Given a sampled command  $c = ([a, ...], \{p, ...\}, \{(v, \tau), ...\})$, we mapped each action $a$ to a set of goal predicates $\{p_a, ...\}$ that it achieves. By combining these predicates with the qualifier predicates $\{p, ...\}$, then adding an existential quantifier for any variables, we created a PDDL goal formula $g_c$ that could be used by a planning algorithm. For example, the command {\small\texttt{(handover robot human ?key1) where (iscolor ?key1 blue)}} would be converted to the goal formula {\small\texttt{(exists (?key1 - key) (and (pickedup-by robot ?key1) (has human ?key1) (iscolor ?key1 blue)))}}.

Some commands had multiple possible groundings. In mDKG (but not in VH), these groundings corresponded to plans with different lengths. As such, we implemented both a \emph{naive} and \emph{efficient} form of literal assistance. In naive assistance, we generated plans for all possible groundings of the lifted command $c$, and averaged performance metrics across these plans. For example, if \texttt{key2} was blue, then a grounding of {\small\texttt{(handover robot human ?key1) where (iscolor ?key1 blue)}} might be {\small\texttt{(handover robot human key2)}}, which translates to the the goal {\small\texttt{(and (pickedup-by robot key2) (has human key2))}}. In efficient assistance, we simply generated a plan for existential goal formula implied by the lifted command, automatically finding the most efficient way to satisfy that formula.

\paragraph{Plan Generation} We generated an assistive plan in two phases. In the first phase, we generated an optimal \emph{joint} plan to the commanded goal $g_c$ using A* search with a node budget of $2^{16}$ (for commands with \texttt{handover} actions, this might involve both agents walking towards a convenient meeting point). If finding a joint plan was unsuccessful, we fell back to computing a plan where only the assistant was allowed to move. In the second phase, we then enforced that only the principal was allowed to move (in mDKG) or pick up items (in VH), and computed the remaining plan to achieve the principal's true goal $g$ (again, using A* search with a node budget of $2^{16}$).

\subsection{Multimodal LLM (GPT-4V)}

We used the public preview of GPT-4V (\texttt{gpt-4-1106-vision-preview}) as a baseline model in our mDKG experiments. We used zero-shot prompting, providing the model with effectively the same instructions as given to the human raters, along with the final frame of the GIF animation shown to humans (Figure \ref{fig:gpt4v-prompt}). Responses were generated with a temperature of 0.

\begin{figure}[b]
\centering
\fbox{\begin{minipage}{0.98\textwidth}
\small
You're watching a human and a robot collaborating on a treasure game shown below. The human player wants to collect exactly one of the four target gems. The two players cooperate to achieve the human's goal.
\vspace{3pt}

During the game, the human may communicate with the robot to coordinate their actions. Any communication that happens is shown below the visuals. 
\vspace{3pt}

The rules of the game are as follows:
\begin{itemize}
    \item The human player's goal is to collect one target gem.
    \item The robot plays an assistive role, and cannot pick up gems on its own.
    \item The players can move on the white squares.
    \item The players have a full view of the map at all times.
    \item  Keys are used to unlock doors with the same color (e.g. red keys only unlock red doors).
    \item Each key can only be used once. Keys disappear after being used.
    \item One player can pass keys to the other if they're next to each other.
    \item Players can occupy the same square and walk past each other.
    \item The keys and doors are labeled. The labels are shown on the top right corner of each cell.
\end{itemize}
\vspace{3pt}

\fbox{stimulus.png}
\vspace{3pt}

In the scenario above, the human player took several actions, then gave the instruction: "Can you hand me that key?"
\vspace{3pt}

Given this information, what gem is most likely to be the human agent's goal? Which key(s) should the robot pick up to best assist the human? 
\vspace{3pt}

Please respond in the following JSON format:

\{
"goal": [red/yellow/blue/green],
"assist": [key1, key2, etc.]
\}
\end{minipage}}
\caption{GPT-4V zero-shot prompt \textmd{for an example key assistance problems in Doors, Keys \& Gems.}}
\label{fig:gpt4v-prompt}
\end{figure}

\section{Human Experiments}
\setcounter{figure}{0}
\setcounter{table}{0}

\begin{figure}[h]
    \centering
    \begin{subfigure}[b]{0.45\textwidth}
    \includegraphics[width=\textwidth]{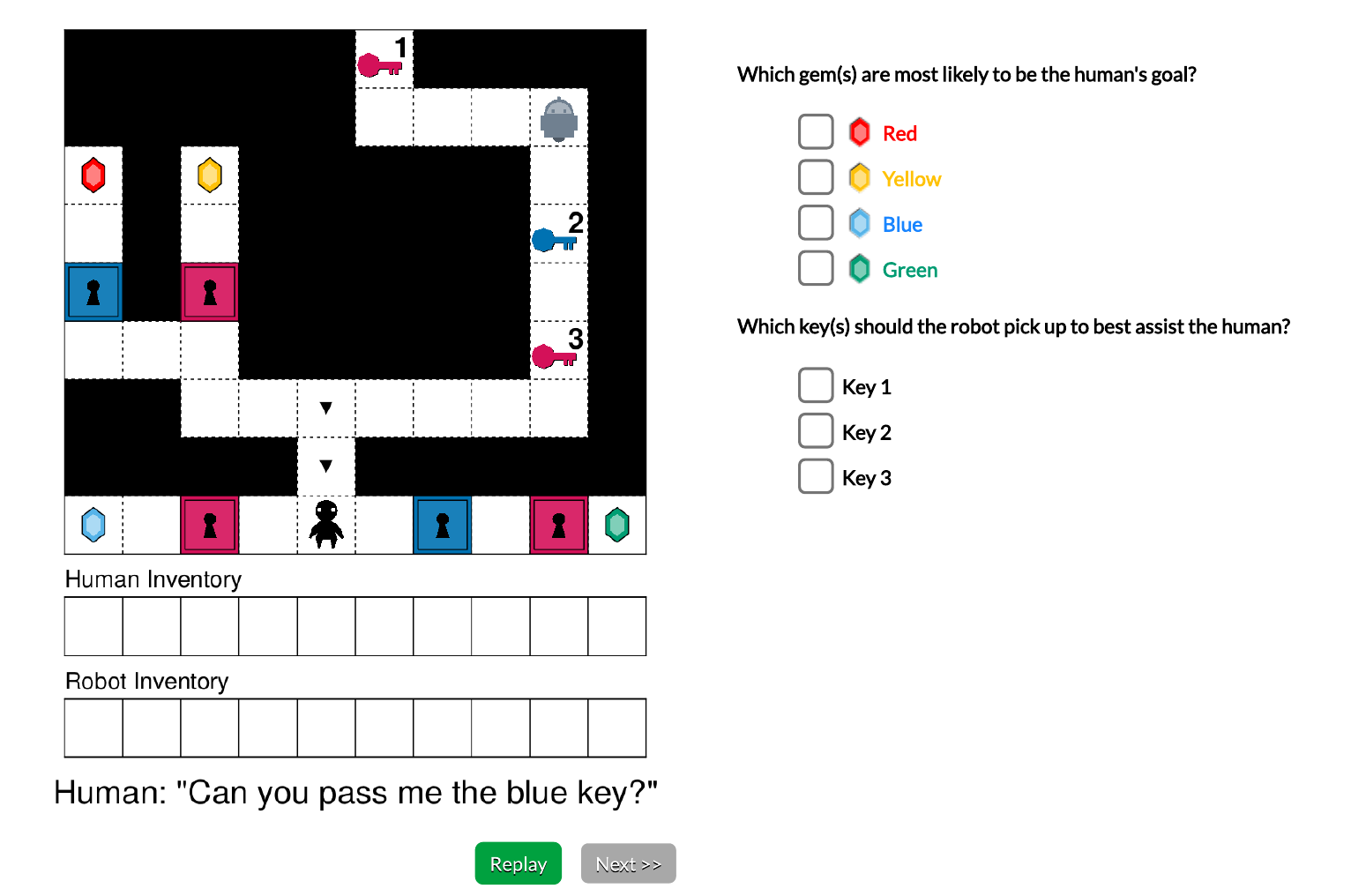}
    \caption{\textmd{Key assistance problem.}}
    \end{subfigure}
    \begin{subfigure}[b]{0.45\textwidth}
    \includegraphics[width=\textwidth]{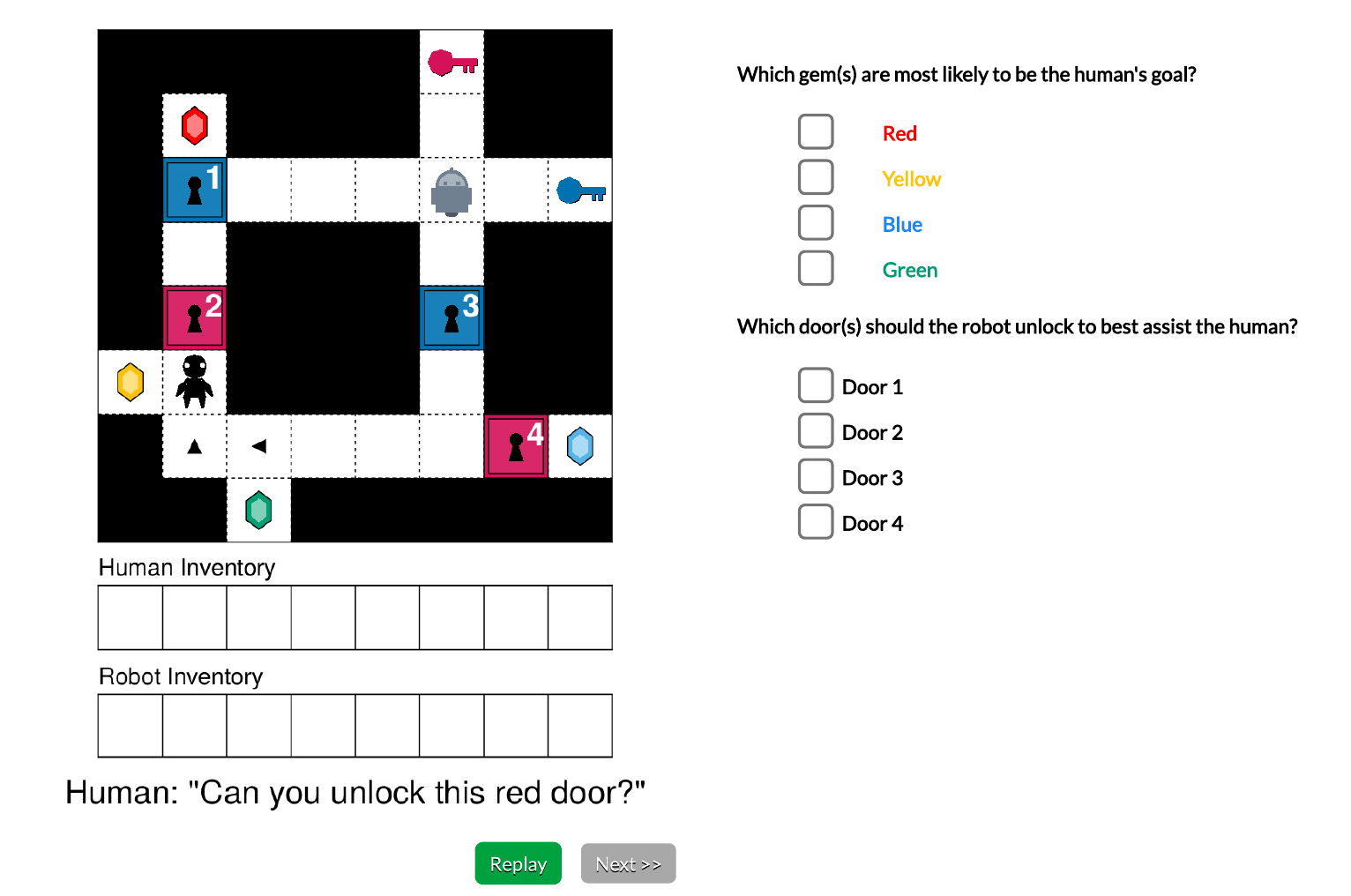}
    \caption{\textmd{Door assistance problem.}}
    \end{subfigure}
    \caption{Web interface for human experiments. \textmd{Participants had to select checkboxes for all gems they thought were likely to be human principal's goal, and select all keys or doors they thought the robot assistant should pick-up or unlock.}}
    \label{fig:interface}
\end{figure}

Here we provide additional details about our human experiments for the mDKG domain. Participants first completed a tutorial that explained the task and experimental interface, then answered 5 comprehension questions before proceeding to the main experiment. In the main experiment, they were shown 15 stimuli in randomized order, which involved either keys or doors as assistance options (Figure \ref{fig:interface}). 

To incentivize accurate but calibrated responses, participants were rewarded for accurately guessing the true goal or selecting the optimal assistance option. Specifically, they earned $1/N$ bonus points if they selected $N$ goals out of which one was the true goal, but $0$ points if none of their selected goals was the true goal. They earned one additional point if they selected exactly the right assistance options. Participants were paid US\$1 for every 30 bonus points they earned, on top of a base pay of US\$16/hr (US\$4 per experiment).

\begin{figure}[h]
\begin{subfigure}[b]{\textwidth}
\includegraphics[width=0.95\textwidth]{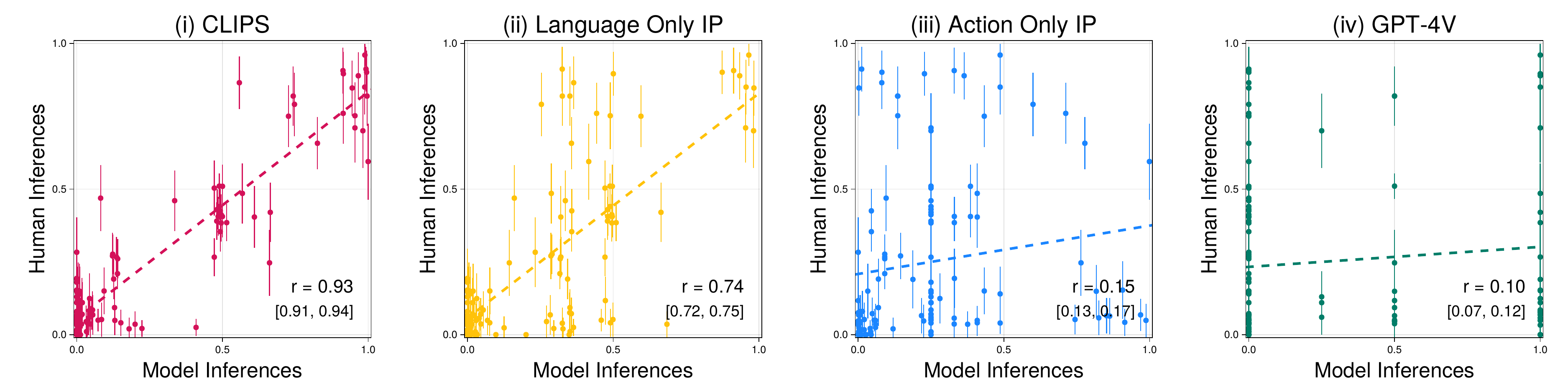}
\caption{Goal Inference: \textmd{Human ($y$-axis) vs. model ($x$-axis) goal inferences.}}
\end{subfigure}
\begin{subfigure}[b]{\textwidth}
\includegraphics[width=0.95\textwidth]{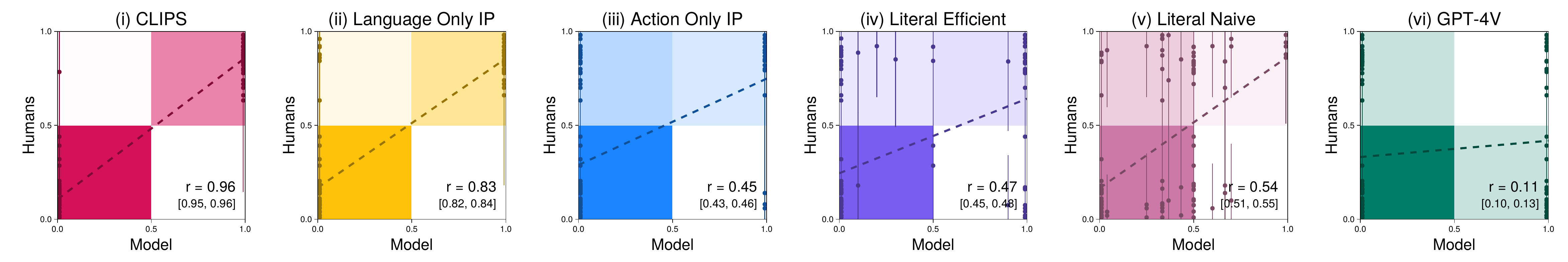}
\caption{Assistance: \textmd{Human ($y$-axis) vs. model ($x$-axis) assistance judgments.}}
\end{subfigure}
\caption{Human-model correlation plots for Doors, Keys \& Gems. \textmd{In \textbf{(a)}, we show scatter plots comparing average human goal inferences vs. model goal inferences across all stimuli. In (b), we compare average human assistance judgments vs. model outputs by overlaying a scatter plot over a binarized confusion matrix. Error bars correspond to 95\% confidence intervals for the mean human response. At the bottom-right of each plot we report Pearson's $r$ (95\% bootstrap CI below).}}
\label{fig:correlation}
\end{figure}

In Figure \ref{fig:correlation}, we show correlation plots that provide a more comprehensive view of how well each model matches human responses. As Figure \ref{fig:correlation}a shows, CLIPS best reproduces human goal inferences. In contrast, unimodal inverse planning tends to be under-confident relative to humans, since they only incorporate evidence from one modality rather than two. GPT-4V tends to be over-confident on stimuli where humans are not (and vice versa), resulting in a very low correlation overall.

To compare human assistance judgments with model outputs, Figure \ref{fig:correlation}b shows scatter plots laid over binarized confusion matrices, generated by thresholding human and model responses at 0.5, then plotting bin counts as a heatmap. Since many of our models return either 1 or 0 for assistance, this provides a better way to visualize human-model agreement. CLIPS best matches human judgments, followed by language only inverse planning. The other models often fail to select an assistance option in cases where humans do (deeper colored top-left quadrants). GPT-4V selects options in many cases where humans do not (bottom-left quadrant), achieving the lowest correlation.

\section{GPT-4V Responses}
\setcounter{figure}{0}
\setcounter{table}{0}

\begin{figure}[h]
\begin{subfigure}[b]{\textwidth}
\begin{minipage}{0.23\textwidth}
\caption{\textmd{Ambiguous predicates.}}
\includegraphics[width=0.95\textwidth]{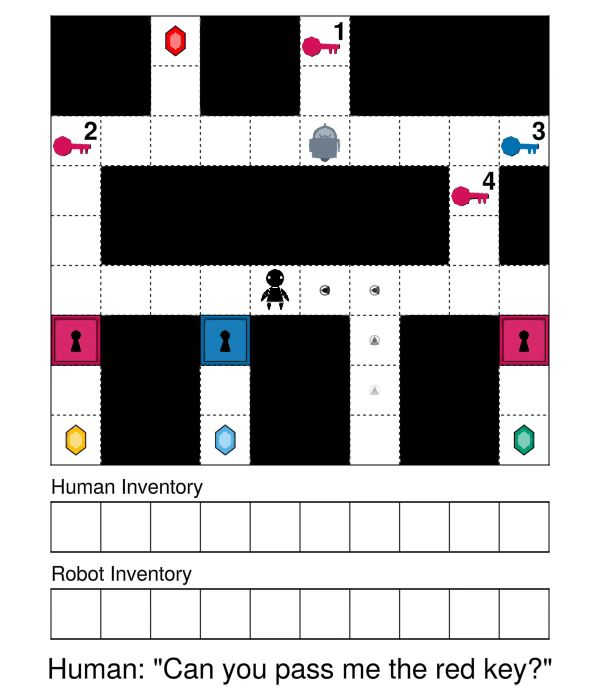}
\end{minipage} \hfill
\fbox{
\begin{minipage}{0.73\textwidth}
\textbf{GPT4V Response:}\\
Based on the image and the rules provided, the human player is asking for the red key, which suggests that their goal is to collect the red gem. The red gem is behind the red door labeled "1," and the red key is required to unlock it. Since the human player can only collect one target gem, and they are asking for the red key, it is most likely that the red gem is their goal.
\\\\
The robot should pick up the red key to assist the human player. The robot cannot pick up gems, so it only needs to focus on the keys that will help the human player achieve their goal.
\\\\
Here is the response in the requested JSON format:

\{
  "goal": ["red"],
  "assist": ["key1"]
\}
\end{minipage}
}
\end{subfigure}
\vspace{6pt}

\begin{subfigure}[b]{\textwidth}
\begin{minipage}{0.23\textwidth}
\caption{\textmd{Ambiguous number.}}
\includegraphics[width=0.95\textwidth]{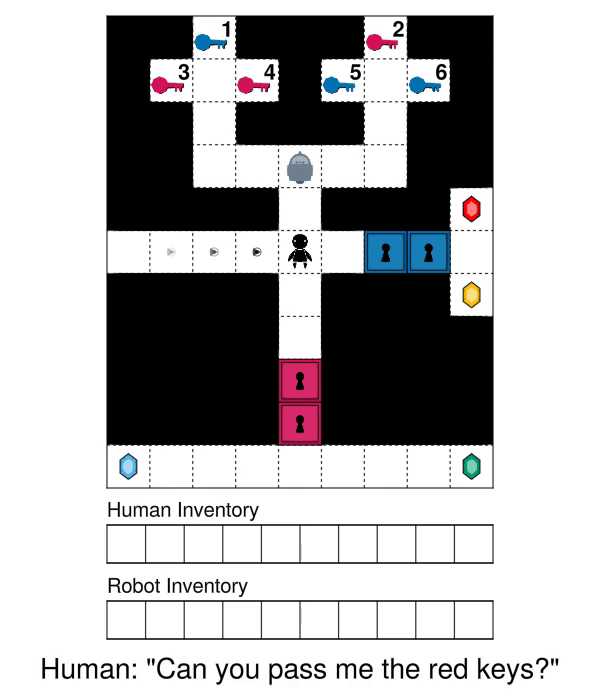}
\end{minipage} \hfill
\fbox{
\begin{minipage}{0.73\textwidth}
\textbf{GPT4V Response:}\\
Based on the image and the information provided, the human player has asked the robot to pass the red keys. This implies that the human player is aiming for a gem that requires passing through a red door. In the image, the red gem is behind a red door, so it is the most likely target for the human player.
\\\\
The robot should pick up the keys that will allow the human to reach the red gem. The red gem is behind door 3, which requires key 3. However, to reach key 3, the robot first needs to unlock door 1 with key 1. Therefore, the robot should pick up key 1 and then key 3 to assist the human player.
\\\\
Here is the response in the requested JSON format:
\\
\{
  "goal": ["red"],
  "assist": ["key1", "key3"]
\}
\end{minipage}
}
\end{subfigure}
\vspace{6pt}

\begin{subfigure}[b]{\textwidth}
\begin{minipage}{0.23\textwidth}
\caption{\textmd{Joint instructions.}}
\includegraphics[width=0.95\textwidth]{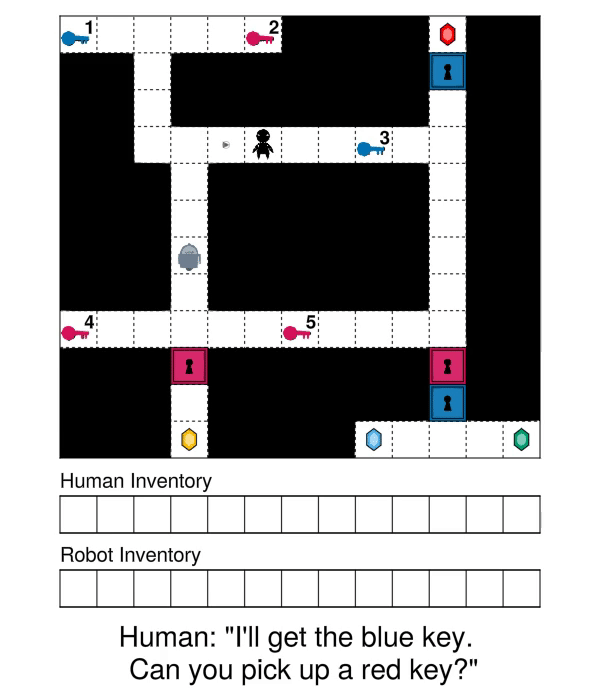}
\end{minipage} \hfill
\fbox{
\begin{minipage}{0.73\textwidth}
\textbf{GPT4V Response:}\\
Based on the information provided and the layout of the game board, the human player has expressed an intention to pick up the blue key. This key can unlock the blue door, which is blocking the path to the green gem. Since the human player's goal is to collect one target gem and they are going for the blue key, it is reasonable to assume that the green gem is their intended target.
\\\\
The robot, in an assistive role, has been asked to pick up a red key. There are two red keys (key4 and key5) and two red doors on the map. However, since the human is going for the green gem, the robot should pick up the red key that will assist in this goal. Key4 is the red key that unlocks the red door on the path to the green gem. Therefore, the robot should pick up key4.
\\\\
Here is the JSON response with the likely goal and the key the robot should pick up:
\\
\{
  "goal": ["green"],
  "assist": ["key4"]
\}
\end{minipage}
}
\end{subfigure}

\caption{Example responses by GPT-4V. \textmd{GPT-4V produces two paragraphs of reasoning text, then gives its final response in JSON.}} 
\label{fig:gpt4v-responses}
\end{figure}

To better illustrate the capabilities and limitations of multimodal LLMs, we show several example responses by GPT-4V in Figure \ref{fig:gpt4v-responses}.

As can be seen from Figure \ref{fig:gpt4v-responses}, GPT-4V is highly inaccurate in its ability to perceive objects in the provided image and determine their spatial relationships, sometimes confabulating object labels that do not exist (Fig. \ref{fig:gpt4v-responses}a--b), failing to notice which gems are locked behind which doors (Fig. \ref{fig:gpt4v-responses}b--c),  and failing to perceive which items are along a path to the goal (Fig. \ref{fig:gpt4v-responses}c). This weakness at spatial perception makes current multimodal LLMs inadequate for assistive tasks like ours, which require coherent spatial reasoning.

However, the reasoning text produced by GPT-4V appears to demonstrate internal means-ends consistency (keys are required to unlock doors of the same color, which in turn allow the human to reach the gems behind them). As prior work has shown \cite{sclar-etal-2023-minding}, this suggests that if an abstract symbolic representation is provided to a state-of-the-art LLM with the appropriate high-level relations (spatial connectivity etc.), they may be able to perform significantly better. Nonetheless, deriving an abstract symbolic representation from the underlying gridworld is itself a form of complex spatial reasoning \cite{aversa2016pruning}, which likely presents a challenge to contemporary LLMs.

A full set of GPT-4V's responses can be found in the Supplementary Information (\href{https://osf.io/v8ru7/}{\texttt{https://osf.io/v8ru7/}}).

\section{Alternate Assistance Methods}
\setcounter{figure}{0}
\setcounter{table}{0}

In the main text, we reported results for an \emph{offline} version of the $Q_\text{MDP}$ assistance policy in Algorithm \ref{alg:clips-assistance}, which keeps the goal posterior fixed after observing $a_{1:T}^h$ and $u_{1:T}$, rather than continuing to update the posterior as the human principal takes more actions. Here we report results for the \emph{online} version of $Q_\text{MDP}$ assistance. We also report results for an alternative assistance method, $\bar\pi_\text{MDP}$, which is constructed by averaging over \emph{policies} $\pi$ for each inferred goal $g$: Rather than minimizing the expected $Q$-value as in $Q_\text{MDP}$ assistance \cite{littman1995learning}, $\bar\pi_\text{MDP}$ first samples a goal $g$ from the posterior, then follows the policy $\pi$ for that goal. This can be viewed as Thompson sampling \cite{thompson1935theory} over policies.

\begin{figure}[h]
    \begin{subfigure}[t]{0.47\textwidth}
    \includegraphics[width = 8.2cm]{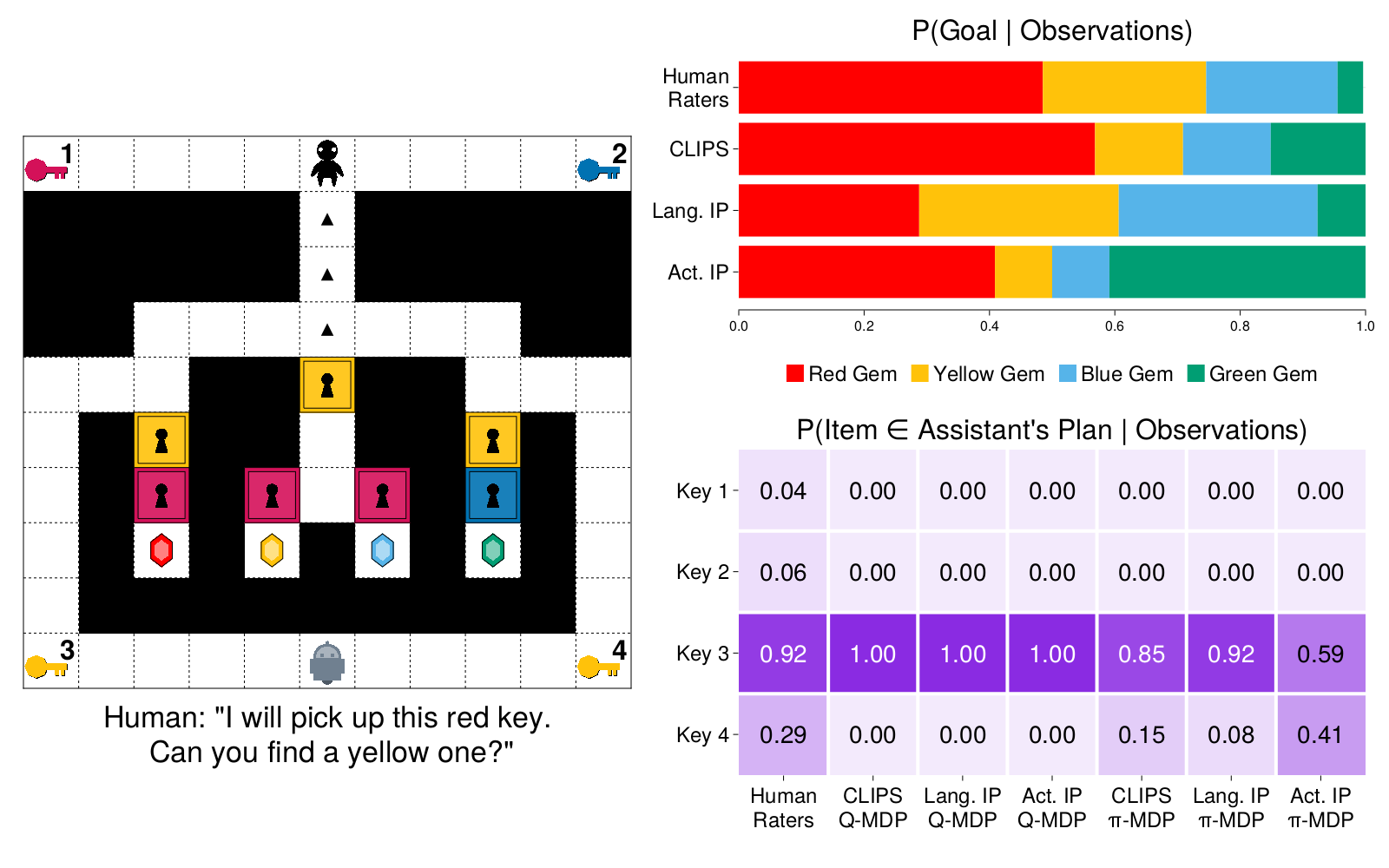}
    \caption{Uncertain Goals. \textmd{When goal uncertainty is high, $Q_\text{MDP}$ assistance acts confidently to to minimize expected goal achievement cost, unlike $\bar\pi_\text{MDP}$ assistance, which is less confident.}}
    \end{subfigure}
    \hspace{0.02\textwidth}
    \begin{subfigure}[t]{0.47\textwidth}
    \includegraphics[width = 8.2cm]{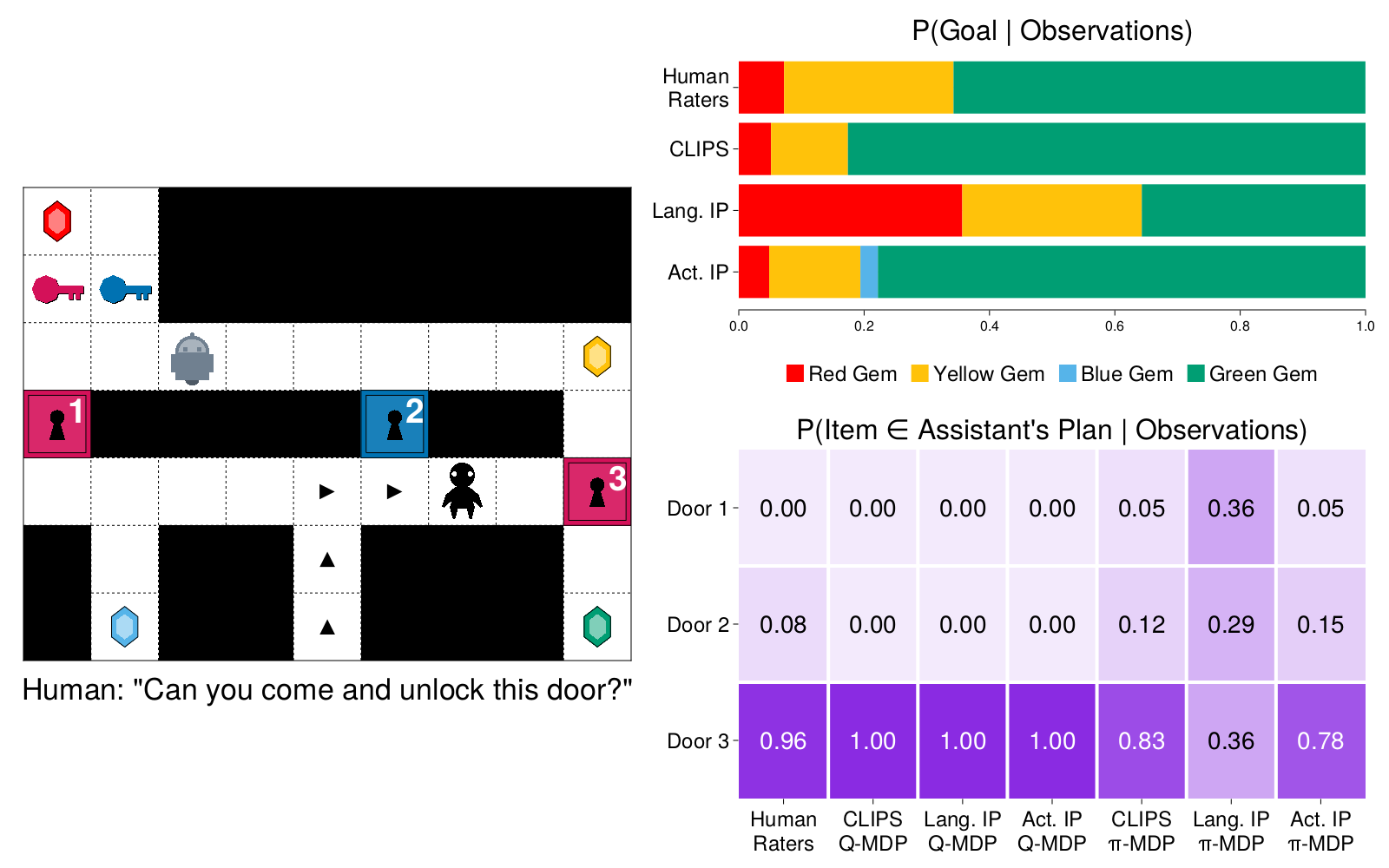}
    \caption{Ambiguous Indexicals. \textmd{Given uncertainty over the goal, $\bar\pi_\text{MDP}$ randomizes over which door is most efficient to unlock. This leads to a higher chance of failure to reach the green gem.}}
    \end{subfigure}
    
    \begin{subfigure}[t]{0.47\textwidth}
    \includegraphics[width = 8.2cm]{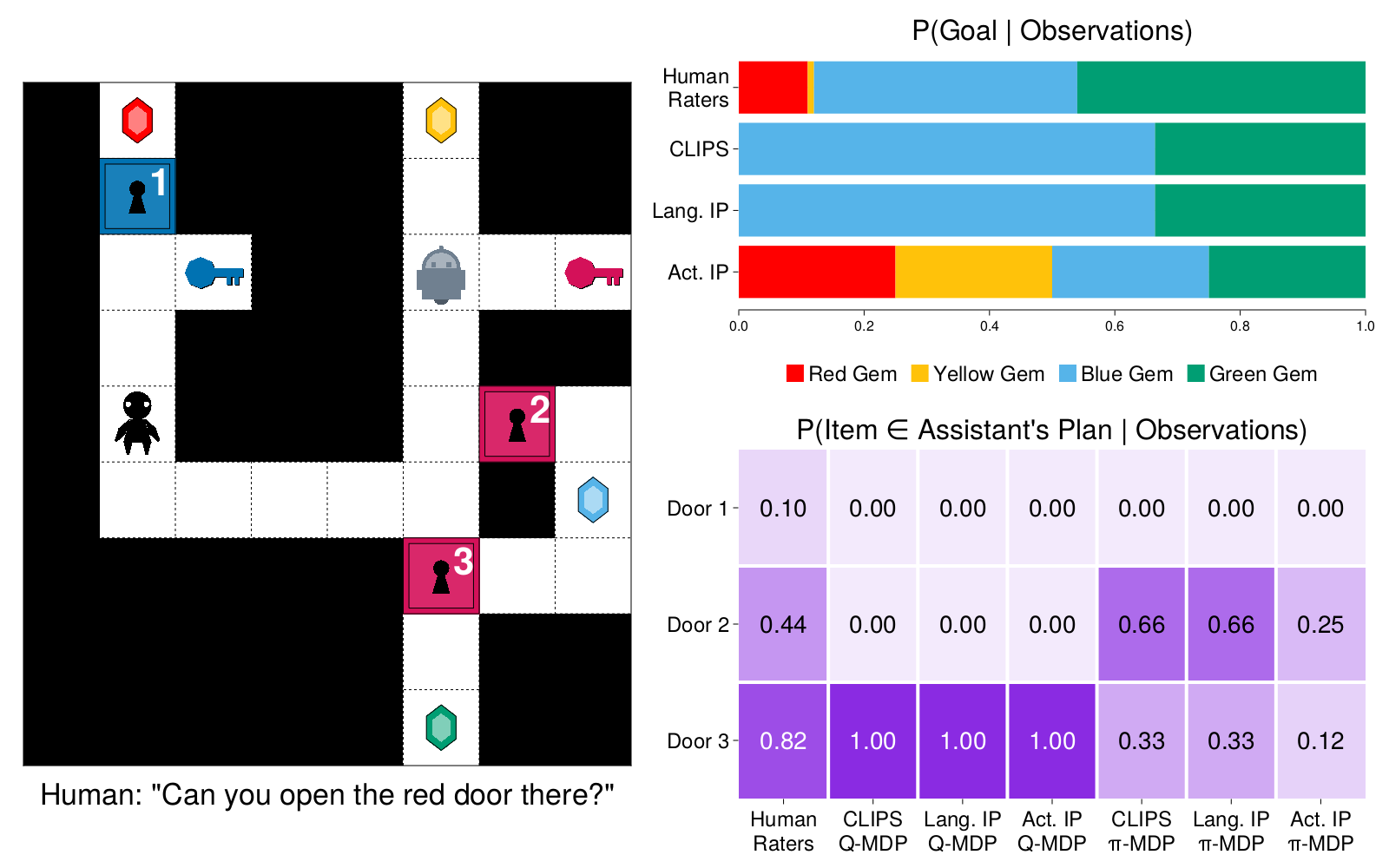}
    \caption{Safe Assistance. \textmd{When the most efficient plan for the most likely goal causes another goal to be unreachable, $\bar\pi_\text{MDP}$ is less safe than $Q_\text{MDP}$ assistance, which acts to ensure that both likely goals are reachable.}}
    \end{subfigure}
    \hspace{0.02\textwidth}
    \begin{subfigure}[t]{0.47\textwidth}
    \includegraphics[width = 8.2cm]{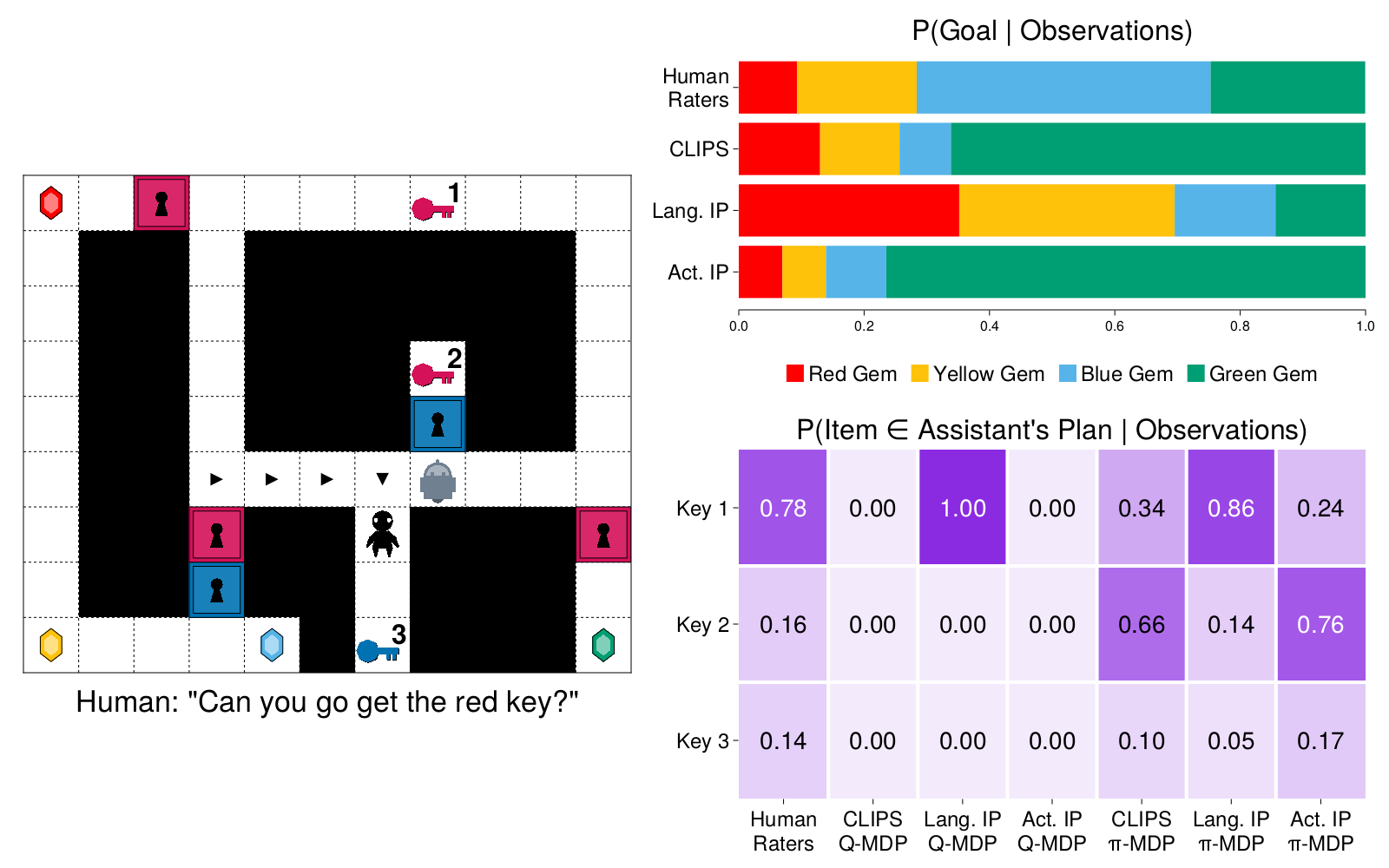}
    \caption{Failed Assistance. \textmd{In the one mDKG problem where CLIPS produces inaccurate goal inferences, $Q_\text{MDP}$ assistance fails, whereas $\bar\pi_\text{MDP}$ sometimes assists the human appropriately by randomizing over goals.}}
    \end{subfigure}

    \caption{Examples comparing (offline) $Q_\textmd{MDP}$ and $\bar\pi_\textmd{MDP}$ assistance in Doors, Keys \& Gems. \textmd{In $Q_\textmd{MDP}$ assistance, we repeatedly take the best action that minimizes the expected goal achievement cost with respect to the posterior over goal specifications. In $\bar\pi_\textmd{MDP}$ assistance, we randomly select a single goal specification $g$ from the posterior, and follow the optimal joint policy for that goal.}}
    \label{fig:dkg-assistance}
\end{figure}

\begin{table}[t]
\centering
\caption{Comparison of assistance methods \textmd{measured in terms accuracy (precision and recall for assistance options), helpfulness (plan length and human cost relative to CLIPS with offline $Q_\text{MDP}$ assistance), and human similarity (correlation of assistance options with mean human ratings) Metrics are averaged across the dataset per domain, with standard errors reported in brackets.}}
\begin{tabular}{llrrrrr}
\hline
 & \multicolumn{1}{c}{\textbf{}} & \multicolumn{2}{c}{\textbf{Accuracy}} & \multicolumn{2}{c}{\textbf{Helpfulness}} & \multicolumn{1}{c}{\textbf{Human Sim.}} \\
\multicolumn{1}{c}{\textbf{Inference}} & \multicolumn{1}{c}{\textbf{Assistance}} & \multicolumn{1}{l}{Assist. Prec.} & \multicolumn{1}{l}{Assist. Rec.} & \multicolumn{1}{l}{Rel. Plan Length} & \multicolumn{1}{l}{Rel. Human Cost} & \multicolumn{1}{l}{Assist. Cor.} \\ \hline
\multicolumn{7}{c}{\textit{Doors, Keys \& Gems}} \\ \hline 
CLIPS & $Q_\text{MDP}$ (Offline) & 0.97 (0.03) & 0.97 (0.03) & 1.00 (0.00) & 1.00 (0.00) & 0.96 (0.01) \\
Lang. Only IP & $Q_\text{MDP}$ (Offline) & 0.90 (0.06) & 0.83 (0.06) & 1.26 (0.10) & 1.18 (0.07) & 0.83 (0.01) \\
Action Only IP & $Q_\text{MDP}$ (Offline) & 0.43 (0.09) & 0.40 (0.09) & 1.68 (0.15) & 1.46 (0.08) & 0.45 (0.01) \vspace{3pt}\\
CLIPS & $Q_\text{MDP}$ (Online) & \textbf{1.00 (0.00)} & \textbf{1.00 (0.00)} & \textbf{0.99 (0.01)} & \textbf{0.99 (0.00)} & \textbf{0.97 (0.00)} \\
Lang. Only IP & $Q_\text{MDP}$ (Online) & 0.90 (0.06) & 0.83 (0.06) & 1.26 (0.10) & 1.18 (0.07) & 0.83 (0.01) \\
Action Only IP & $Q_\text{MDP}$ (Online) & 0.88 (0.06) & 0.90 (0.06) & 1.15 (0.07) & 1.09 (0.03) & 0.83 (0.01) \vspace{3pt} \\
CLIPS & $\bar\pi_\text{MDP}$ & 0.91 (0.03) & 0.91 (0.03) & 1.09 (0.04) & 1.08 (0.02) & 0.96 (0.00) \\
Lang. Only IP & $\bar\pi_\text{MDP}$ & 0.79 (0.05) & 0.77 (0.05) & 1.31 (0.09) & 1.19 (0.04) & 0.85 (0.01) \\
Action Only IP & $\bar\pi_\text{MDP}$ & 0.72 (0.04) & 0.41 (0.05) & 1.58 (0.12) & 1.45 (0.05) & 0.58 (0.01) \vspace{3pt} \\ \hline
\multicolumn{7}{c}{\textit{VirtualHome}} \\ \hline
CLIPS & $Q_\text{MDP}$ (Offline) & 0.87 (0.04) & \textbf{1.00 (0.00)} & \textbf{1.00 (0.00)} & \textbf{1.00 (0.00)} & \textbf{--} \\
Lang. Only IP & $Q_\text{MDP}$ (Offline) & 0.59 (0.05) & 0.96 (0.03) & 1.33 (0.07) & 1.35 (0.08) & \textbf{--} \\
Action Only IP & $Q_\text{MDP}$ (Offline) & 0.61 (0.07) & 0.84 (0.07) & 1.30 (0.07) & 1.33 (0.07) & \textbf{--} \vspace{3pt} \\
CLIPS & $Q_\text{MDP}$ (Online) & \textbf{0.90 (0.04)} & \textbf{1.00 (0.00)} & 1.12 (0.15) & 1.13 (0.17) & \textbf{--} \\
Lang. Only IP & $Q_\text{MDP}$ (Online) & 0.59 (0.05) & 0.96 (0.03) & 1.33 (0.07) & 1.35 (0.08) & \textbf{--} \\
Action Only IP & $Q_\text{MDP}$ (Online) & 0.84 (0.06) & 0.92 (0.06) & 1.03 (0.04) & 1.04 (0.05) & \textbf{--} \vspace{3pt} \\
CLIPS & $\bar\pi_\text{MDP}$ & 0.59 (0.06) & 0.65 (0.06) & 1.22 (0.06) & 1.33 (0.07) & \textbf{--} \\
Lang. Only IP & $\bar\pi_\text{MDP}$ & 0.45 (0.05) & 0.57 (0.06) & 1.37 (0.08) & 1.48 (0.09) & \textbf{--} \\
Action Only IP & $\bar\pi_\text{MDP}$ & 0.46 (0.06) & 0.45 (0.05) & 1.61 (0.14) & 1.77 (0.16) & \textbf{--} \vspace{3pt} \\ \hline
\end{tabular}
\label{tab:assistance-results}
\end{table}

\paragraph{Qualitative analysis} Figure \ref{fig:dkg-assistance} illustrates how (offline) $Q_\text{MDP}$ assistance differs from $\bar\pi_\text{MDP}$. While $\bar\pi_\text{MDP}$ assistance better reproduces the gradedness seen in human assistance judgments, this comes at the cost of reduced safety and helpfulness. By following a policy $\pi$ to goal $g$ according to how likely $g$ is, $\bar\pi_\text{MDP}$ fails to aggregate information across multiple uncertain goals when deciding the best action to take. This is most clearly illustrated in Figure \ref{fig:dkg-assistance}c, where CLIPS assigns higher probability to the blue gem over the green gem. $Q_\text{MDP}$ assistance safely unlocks \emph{Door 3} instead of \emph{Door 2}, since doing so will ensure that both the blue gem and green gem are reachable. In contrast, $\bar\pi_\text{MDP}$ unlocks \emph{Door 2} on average 66\% of the time, which is unsafe. However, as Figure \ref{fig:dkg-assistance}d shows, $\bar\pi_\text{MDP}$ is slightly better when CLIPS produces the wrong goal inferences (due to the sub-optimality of the human's actions with respect to the true goal). In this single failure case, $Q_\text{MDP}$ confidently executes the wrong actions, whereas $\bar\pi_\text{MDP}$ assistance is sometimes successful.

\paragraph{Quantitative analysis} We present the performance of different assistance methods in Table \ref{tab:assistance-results}. Compared to offline $Q_\text{MDP}$ assistance, $\bar\pi_\text{MDP}$ assistance was less accurate, and produced longer plans on average. This highlights the importance of minimizing expected cost under uncertainty, rather than just sampling a likely goal to follow. As might be expected, online $Q_\text{MDP}$ assistance was more accurate, since the posterior over goals was updated as new actions were observed. This significantly improved the performance of action-only inverse planning (note that language-only inverse planning is the same regardless of assistance method, since the human principal communicates no further instructions). However, multimodal goal inference via CLIPS combined with online $Q_\text{MDP}$ assistance was still the most accurate, since it leveraged linguistic information from the beginning of assistance.

Apart from one failure case in VirtualHome where the assistant and simulated human principal were caught in a livelock, CLIPS with online $Q_\text{MDP}$ assistance was slightly more helpful on average than offline assistance. Nonetheless, offline assistance was almost as helpful. This suggests that once enough actions are observed to disambiguate the principal's instruction, CLIPS can still assist effectively even if it has partial to no observability of the principal's subsequent actions.

\end{document}